\documentclass[runningheads]{llncs}
\usepackage[utf8]{inputenc}
\usepackage{graphicx}
\usepackage[misc]{ifsym}
\usepackage{url}
\usepackage[utf8]{inputenc}
\usepackage[english]{babel}
\usepackage{bbding}
\usepackage{rotating}
\usepackage{verbatim}
\usepackage{multirow}
\usepackage{algorithm} 
\usepackage{amsmath}
\usepackage{algpseudocode}

\usepackage{amsthm}
\theoremstyle{definition}
\usepackage[flushleft]{threeparttable}
\usepackage{colortbl}
\usepackage[table]{xcolor}
\usepackage{lipsum}
\usepackage{amssymb}
\usepackage{subfigure}
\algnewcommand\algorithmicforeach{\textbf{for each}}
\algdef{S}[FOR]{ForEach}[1]{\algorithmicforeach\ #1\ \algorithmicdo}
\newtheorem{mydef}{Definition}
\newcommand{\el}{{\cal {EL}}}
\title{Repairing $\el$ Ontologies \\ Using Weakening and Completing}

\author{Ying Li \inst{1} \orcidID{0000-0002-1367-9679} \and
Patrick Lambrix \inst{1,2} \Letter \orcidID{0000-0002-9084-0470}}

\institute{
Link{\"o}ping University, Sweden  \\
\and
University of G{\"a}vle, Sweden\\
\email{firstname.lastname@liu.se} 
}

\begin{document}
\maketitle

\begin{abstract}
The quality of ontologies in terms of their correctness and completeness is crucial for developing high-quality ontology-based applications. 
Traditional debugging techniques repair ontologies by removing unwanted axioms, but may thereby remove consequences that are correct in the domain of the ontology. In this paper we propose an interactive approach to mitigate this for $\el$ ontologies by axiom weakening and completing. We present algorithms for weakening and completing and present the first approach for repairing that takes into account removing, weakening and completing. We show different combination strategies, discuss the influence on the final ontologies and show experimental results. We show that previous work has only considered special cases and that there is a trade-off between the amount of validation work for a domain expert and the quality of the ontology in terms of correctness and completeness.
\end{abstract}

\section{Introduction}

Debugging ontologies aims to remove unwanted knowledge in the ontology. This can be knowledge that leads to logical problems such as inconsistency or incoherence (semantic defects) or statements that are not correct in the domain of the ontology (modeling defects) (e.g., \cite{KPSH05}). The workflow consists of several steps including the detection and localization of the defects and the repairing. In this paper we assume we have detected and localized the defects, e.g., using traditional debugging techniques as in, e.g., 
\cite{SC03,Sch05,SHCvH07,MLBP06,KPSH05,KPSC06,KPHS07,LB10,JGHZ11,MMV11,SFFR12,FMVN13,AMIMPM16,RS18,SRS18},   
and we now need to repair the ontology. 
In the classical approaches for debugging the end result is a set of axioms to remove from the ontology that is obtained after detection and localization, and the repairing consists solely of removing the suggested axioms. However, first, these approaches are usually purely logic-based and therefore may suggest to remove correct axioms (e.g., \cite{PFSC13}). Therefore, it is argued that a domain expert should validate the results of such systems. 
Furthermore, removing an axiom may remove more knowledge than necessary. It may happen that correct knowledge that is derivable with the help of the wrong axioms will not be derivable in the new ontology. 
In this paper we mitigate these effects of removing wrong axioms by, in addition to removing those axioms, also adding correct knowledge. 
Two approaches could be used. A first approach is to replace an axiom with a weakened version of the axiom (e.g., \cite{LSPV08,DQF14,BKNP18,TCGPPK18}).
Another approach is to complete an ontology (e.g., \cite{WDL14}) which adds previously unknown correct axioms that allow to derive existing axioms, and that could be used on the results of weakening. These approaches have, however, not been studied together.

In this paper we focus on $\el$ ontologies. $\el$ is a description logic for which subsumption checking remains tractable and that is used (as is or with small extensions) by well-known ontologies such as SNOMED or Gene Ontology \cite{BBL05}. Further, we assume that we are given a set of wrong axioms $W$ that we want to remove from the ontology and that when removing these axioms, they cannot be derived from the ontology anymore. Our contributions are the following: (i) In section \ref{sec-problem-formulation} we formally define the repairing problem. 
(ii) In section \ref{sec-strategies} we define algorithms for weakening and completing and their combinations.  For 
weakening 
we restrict the search space to obtain an efficient algorithm.
The algorithm for completing is an extension of the approach in \cite{WDL14}. Further, to our knowledge, we introduce the first approach for repairing that takes into account removing, weakening and completing. (iii) In section \ref{sec-experiments} we show results of experiments with different ways of combining removing, weakening and completing, as well as the approach in general. Further, we discuss the results and validation issues in section \ref{sec-discussion}. We note that previous work on debugging and weakening (knowingly or not) only used one of the possible combinations that we introduce here. Our implemented systems are presented in section \ref{sec-systems}.
Related work is discussed in section \ref{sec-related-work}, and we conclude in section \ref{sec-conclusion}. We start with preliminaries in section \ref{sec-preliminaries}.

\section{Preliminaries}
\label{sec-preliminaries}

In this paper we assume that ontologies are represented using a description logic TBox. Description logics \cite{BCMNP03} are knowledge representation languages. In description logics, concept descriptions are constructed inductively from a set $N_C$ of atomic concepts and a set $N_R$ of atomic roles and (possibly) a set $N_I$ of individual names. 
Different description logics allow for different constructors for defining complex concepts and roles. 
An interpretation $\cal I$ consists of a non-empty set $\Delta^{\cal I}$ and an interpretation function $\cdot^{\cal I}$ which assigns to each atomic concept $P \in N_C$ a subset $P^{\cal I} \subseteq \Delta^{\cal I}$, to each atomic role $r \in N_R$ a relation $r^{\cal I} \subseteq \Delta^{\cal I} \times \Delta^{\cal I}$, and to each individual name\footnote{As we do not deal with individuals in this paper, we do not use individuals in the later sections.} $i \in N_I$ an element $i^{\cal I} \in \Delta^{\cal I}$.
The interpretation function is straightforwardly extended to complex concepts. A TBox is a finite set of axioms which in $\el$ are \emph{general concept inclusions} (GCIs). 
The syntax and semantics for $\el$ are shown in Table \ref{tab:el}. 

\begin{table}[hbt]
\begin{center}
\caption{$\el$ syntax and semantics. (Note that $P$ and $Q$ are arbitrary concepts. In the remainder we often use $P$ and $Q$ for atomic concepts.)}
\begin{tabular}{|c|c|c|}

\hline
Name & Syntax & Semantics \\
\hline
\hline
top & $\top$ & $\Delta^{\cal I}$ \\
\hline
conjunction & $P \sqcap Q$ & $P^{\cal I} \cap Q^{\cal I}$ \\
\hline
existential restriction & $\exists r. P$ & $\{ x \in \Delta^{\cal I}$ $\mid$ $\exists y \in \Delta^{\cal I} :  $  $(x,y) \in r^{\cal I} \wedge y \in P^{\cal I}\}$\\
\hline
\hline
GCI & $P \sqsubseteq Q$ & $P^{\cal I} \subseteq Q^{\cal I}$\\
\hline
\end{tabular}
\end{center}
\label{tab:el}
\end{table}

An interpretation ${\cal I}$ is a \emph{model} of a TBox ${\cal T}$ if  for each GCI  in ${\cal T}$, the semantic conditions are satisfied.\footnote{We do not take up consistency of TBoxes, i.e., whether a model exists or not, in this paper as every $\el$ TBox is consistent.} 
One of the main reasoning tasks for description logics is subsumption checking in which the problem is to decide for a TBox ${\cal T}$ and concepts $P$ and $Q$ whether ${\cal T}$  $\models P \sqsubseteq Q$, i.e., whether $P^{\cal I} \subseteq Q^{\cal I}$ for every model of TBox ${\cal T}$. In this paper we update the TBox during the repairing and we always use subsumption with respect to the current TBox.

\section{Problem formulation}
\label{sec-problem-formulation}

We can now formally define the repairing problem that we want to solve (Definition \ref{def-repair-ontology}). We are given a set of wrong axioms $W$ that we want to remove from the ontology and that when they are removed, they cannot be derived from the TBox representing the ontology anymore.\footnote{We note that in this paper we deal with removing axioms with the assumption that when they are removed, they cannot be derived from the TBox representing the ontology anymore. This is not the full debugging problem, for which the combination with weakening and completing is left for future work. Removing can be seen as a simple kind of debugging, or as the second step of the debugging process. As an example, assume a TBox with  axioms \textit{A} $\sqsubseteq$ \textit{B}, \textit{B} $\sqsubseteq$ \textit{C}, and \textit{A} $\sqsubseteq$ \textit{C}. Assume that \textit{A} $\sqsubseteq$ \textit{B} and \textit{A} $\sqsubseteq$ \textit{C} are wrong axioms. In full debugging it would be possible to set $W$ = \{\textit{A} $\sqsubseteq$ \textit{C}\}. The system would then compute that \textit{A} $\sqsubseteq$ \textit{C} as well as one of \textit{A} $\sqsubseteq$ \textit{B} and \textit{B} $\sqsubseteq$ \textit{C} need to be removed. A domain expert may then choose to remove \textit{A} $\sqsubseteq$ \textit{B} and \textit{A} $\sqsubseteq$ \textit{C}. In our problem it is not possible that $W$ = \{\textit{A} $\sqsubseteq$ \textit{C}\} as, when removing \textit{A} $\sqsubseteq$ \textit{C}, it still can be derived from the remaining axioms. We thus assume that a first debugging step has been performed, e.g., using traditional methods, and then start with $W$ = \{\textit{A} $\sqsubseteq$ \textit{C}, \textit{A} $\sqsubseteq$ \textit{B}\}. Combining full debugging, weakening and completing will add additional complexity to the already complex problem we describe in this paper.} 
Further, we assume an oracle (representing a domain expert) that when given an axiom, can answer whether this axiom is correct or wrong in the domain of interest of the ontology. In this work we have not required specific properties regarding the performance of the oracle. For instance, we did not require that an oracle always answers correctly or that the oracle gives consistent answers. As a first step we have chosen this way as it reflects reality. According to our long experience working with domain experts in ontology engineering, domain experts make mistakes. However, this does not necessarily mean that domain expert validation is not useful. In experiments in ontology alignment, it was shown that oracles making up to 30\% mistakes were still beneficial (e.g., \cite{DILL17}). Further, requiring consistent answers seems to be a tough requirement for domain experts. This would require the ability to reason with long proof chains, while humans usually do well for chains of length up to circa 7. It is also not clear how to check that a particular domain expert would fulfil the required properties. Therefore, in this work we do not require such properties, but provide user support in our systems by providing warnings when incompatible validations are made and then allow the domain expert to revise the validations. We do acknowledge, however, that requiring such properties and thereby classifying types of domain experts (e.g., \cite{LWDI13}), may allow us to guarantee certain properties regarding correctness and completeness and allow us to reduce the search space of possible repairs.

A repair for the ontology given the TBox ${\cal T}$, oracle $Or$ and $W$, is a set of correct axioms that when added to the TBox where the axioms in $W$ are removed will not allow deriving the axioms in $W$.\footnote{In the full debugging problem we would, in addition to set $A$ in Definition \ref{def-repair-ontology}, also introduce a set $D$ of axioms to remove, such that all axioms in $D$ are false according to the oracle. Further, requirement (ii) would  be replaced by $\forall$ $\psi$ $\in$ $W$: $({\cal T} \cup A) \setminus D$ $\not \models$ $\psi$. See \cite{lambrix20}.}

\begin{mydef} (Repair)
Let ${\cal T}$ be a TBox. 
Let $Or$ be an oracle that given a TBox axiom returns true or false.
Let $W$ be a finite set of TBox axioms in ${\cal T}$ such that $\forall$ $\psi$ $\in$ $W$: $Or$($\psi$) = false.
Then, a repair for Debug-Problem DP$({\cal T}, Or, W)$ is a finite set of TBox axioms $A$ such that \\
(i) $\forall$ $\psi$ $\in$ $A$: $Or$($\psi$) = true;\\
(ii) $\forall$ $\psi$ $\in$ $W$: $({\cal T} \cup A) \setminus W$ $\not \models$ $\psi$. 
\label{def-repair-ontology}
\end{mydef}

Our aim is to find repairs that remove as much wrong knowledge and add as much correct knowledge to our ontology as possible. Therefore, we introduce the preference relations \textit{less incorrect} and \textit{more complete} between ontologies (Definition \ref{def-preferences-ontologies}) and repairs (Definition \ref{def-preferences-repairs}) that formalize these intuitions, respectively.

\begin{mydef} (less incorrect/more complete - ontologies)
\label{def-preferences-ontologies}
Let ${\cal O}_1$ and ${\cal O}_2$ be two ontologies represented by TBoxes ${\cal T}_1$ and ${\cal T}_2$ respectively. \\
Then, ${\cal O}_1$ is \emph{less incorrect} than  ${\cal O}_2$ (${\cal O}_2$ is \emph{more incorrect} than  ${\cal O}_1$) iff 
$(\forall \psi: ({\cal T}_1 \models \psi \wedge Or(\psi) = false )   \rightarrow  {\cal T}_2 \models \psi)) 
\wedge (\exists \psi:  Or(\psi) = false ~\wedge {\cal T}_1 \not\models \psi  \wedge {\cal T}_2 \models \psi)$.\\
${\cal O}_1$  and ${\cal O}_2$ are \emph{equally incorrect} iff 
$\forall \psi: Or(\psi) = false  \rightarrow 
({\cal T}_1 \models \psi  \leftrightarrow  {\cal T}_2 \models \psi)$ \\
Further,  ${\cal O}_1$ is \emph{more complete} than  ${\cal O}_2$ (or ${\cal O}_2$ is \emph{less complete} than  ${\cal O}_1$) iff 
$(\forall \psi: ({\cal T}_2  \models \psi \wedge ~Or(\psi) = true)    \rightarrow  {\cal T}_1 \models \psi)) 
\wedge (\exists \psi:  Or(\psi) = true ~\wedge {\cal T}_1  \models \psi \wedge {\cal T}_2 \not\models \psi)$.\\
${\cal O}_1$  and ${\cal O}_2$  are \emph{equally complete} iff 
$\forall \psi: Or(\psi) = true  \rightarrow 
({\cal T}_1 \models \psi  \leftrightarrow  {\cal T}_2 \models \psi)$
\end{mydef}

\begin{mydef} (less incorrect/more complete - repairs)
\label{def-preferences-repairs}
Let ${\cal O}$ be an ontology represented by TBox ${\cal T}$ and let $A_1$ and $A_2$ be two repairs for DP$({\cal T}, Or, W)$. Let ${\cal O}_1$ be the ontology represented by $(({\cal T} \cup A_1) \setminus W)$ and ${\cal O}_2$  the ontology represented by $(({\cal T} \cup A_2) \setminus W)$.
Then, $A_1$ is \emph{less incorrect} than $A_2$ (or $A_1$ is preferred to $A_2$ with respect to `less incorrect') iff 
${\cal O}_1$ is \emph{less incorrect} than  ${\cal O}_2$.
Further, repair $A_1$ is \emph{more complete} than repair $A_2$ (or $A_1$ is preferred to $A_2$ with respect to `more complete') iff 
${\cal O}_1$ is \emph{more complete} than  ${\cal O}_2$.
\end{mydef}

\section{Strategies}
\label{sec-strategies}

We now define algorithms for weakening and completing and their combinations. 

\subsection{Basics}

We assume that the TBoxes representing ontologies are  \textit{normalized} $\el$ TBoxes. 
A normalized $\el$ TBox ${\cal T}$ contains only axioms of the forms  $P \sqsubseteq Q$, $P \sqcap Q$ $\sqsubseteq$ $R$, $\exists r. P$ $\sqsubseteq$ $Q$ and $P$ $\sqsubseteq$ $\exists r. Q$ where $P$, $Q$, $R$ $\in$ $N_C$ 
and $r$ $\in$ $N_R$. Every $\el$ TBox can in linear time be transformed into a normalized TBox that is a conservative extension, i.e., every model of the normalized TBox is also a model of the original TBox and every model of the original TBox can be extended to a model of the normalized TBox \cite{BBL05}.

Further, we define the \textit{simple complex concept set for a TBox ${\cal T}$}, which contains all atomic concepts in the ontology as well as the concepts that can be constructed by using one constructor ($\cap$ or $\exists$) and only atomic concepts and roles in the ontology (Definition \ref{def-SCC}). Note that $\top$ is not in SCC(${\cal T}$). Further,
if the number of concepts in $N_C^{\cal T}$ is $n$ and the number of roles in $N_R^{\cal T}$ is $t$, then the number of concepts in SCC(${\cal T}$) is 
$(n^2+n)/2 + tn$.

\begin{mydef} 
For a normalized $\el$ TBox $\cal{T}$ with $N_C^{\cal T}$ the set of atomic concepts occurring in $\cal{T}$ and $N_R^{\cal T}$ the set of atomic roles occurring in $\cal{T}$, 
we define the \textit{simple complex concept set for ${\cal T}$}, denoted by SCC(${\cal T}$), as the set containing 
all the concepts of the forms $P$, $P$ $\sqcap$ $Q$, and $\exists r. P$ where $P$, $Q$ $\in$ $N_C^{\cal T}$
and $r$ $\in$ $N_R^{\cal T}$. 
\label{def-SCC}
\end{mydef}


In our algorithms we use two basic operations which remove and add axioms to a TBox. The result of \textit{Remove-axioms}(${\cal T}$,$D$) for a TBox ${\cal T}$ and a set of axioms $D$ is the TBox ${\cal T}$ $\setminus$ $D$.   If $D$ contains only wrong axioms (such as $W$), then the ontology represented by Remove-axioms(${\cal T}$,$D$) is 
less (if at least one of the removed axioms cannot be derived anymore) or equally incorrect (if all removed axioms can still be derived), 
as well as less (if some correct axioms cannot be derived anymore by removing the wrong ones)  or equally complete (if all correct axioms can still be derived),
than the ontology represented by ${\cal T}$. 
The result of \textit{Add-axioms}(${\cal T}$,$A$) for a TBox ${\cal T}$ and a set of axioms $A$ is the TBox ${\cal T}$ $\cup$ $A$.  
If $A$ contains only correct axioms 
then the ontology represented by Add-axioms(${\cal T}$,$A$) is 
more (if some added axiom was not derivable from the ontology) or equally complete (if all added axioms were derivable from the ontology), 
as well as more (if some wrong axioms can now be derived by adding the new ones) or equally incorrect (if no new wrong axioms can now be derived by adding the new ones),
than the ontology represented by ${\cal T}$. 

We also need to compute sub-concepts and super-concepts of concepts. However, to reduce the infinite search space of possible axioms to add during weakening and completing, we limit the use of nesting operators while computing sub- and super-concepts.\footnote{This limitation allows us to restrict the search space. Weaker limitations are possible, but the weaker the restriction, the larger the search space of possible solutions and the higher the probability of a less usable practical system.} This we do by only considering sub- and super-concepts in the SCC of a TBox (Definition \ref{def-SCC-super-sub-concepts}). As subsumption checking in $\el$ is tractable, finding these sub- and super-concepts is tractable.

\begin{mydef} (super- and sub-concepts in SCC)\\
$sup(P,{\cal T})\leftarrow $\{ $sp$ $\mid$ ${\cal T}$ 
    $\vDash$ 
    $P \sqsubseteq sp$ $\wedge$ $sp \in$ SCC(${\cal T}$)\} \\
$sub(P,{\cal T})\leftarrow$\{ $sb$ $\mid$ ${\cal T}$ 
    $\vDash$ 
    $sb \sqsubseteq P$ $\wedge$ $sb$ $\in$ SCC(${\cal T}$)\}
\label{def-SCC-super-sub-concepts}
\end{mydef}
Finally, as we work on normalized $\el$ TBoxes, we need to make sure that when adding axioms, these are of one of the forms $P \sqsubseteq Q$, $P \sqcap Q$ $\sqsubseteq$ $R$, $\exists r. P$ $\sqsubseteq$ $Q$ and $P$ $\sqsubseteq$ $\exists r. Q$ where $P$, $Q$, $R$ $\in$ $N_C^{\cal T}$  
and $r$ $\in$ $N_R^{\cal T}$ . Algorithm \ref{alg-normalization} rewrites an axiom into one of the allowed forms. We note that 
for some cases (lines 10-15) new atomic concepts, not originally in the ontology, may be introduced.

\begin{algorithm}
	\caption{Normalize($sb \sqsubseteq sp$)} 
	\hspace*{\algorithmicindent}\textbf{Input}: Axiom  $sb \sqsubseteq sp$ \\
   \hspace*{\algorithmicindent}\textbf{Output}: A set of axioms in normalized form
	\begin{algorithmic}[1]
   \If{$sp$ $\in$ ${N_c}$}
   \State \Return \{ $sb \sqsubseteq sp$ \}
   \ElsIf{$sp$ is of the form $P \sqcap Q$}
   \State \Return \{ $sb \sqsubseteq P$, $sb \sqsubseteq Q$ \}
   \ElsIf{$sp$ is of the form $\exists r.P$}
        \If{$sb$ $\in$ ${N_c}$}
          \State  \Return \{ $sb \sqsubseteq sp$ \}
        \ElsIf{$sb$ is of the form $\exists r.Q$}
          \State Introduce new concept $Z$
          \State \Return \{ $\exists r.Q \sqsubseteq Z$, $Z \sqsubseteq \exists r.Q$, $Z \sqsubseteq sp$\}
        \ElsIf{$sb$ is of the form $\exists s.Q$}
           \State Introduce new concept $Z$
          \State \Return \{ $\exists s.Q \sqsubseteq Z$, $Z \sqsubseteq \exists s.Q$, $Z \sqsubseteq sp$\}
        \ElsIf{$sb$ is of the form $Q \sqcap R$} 
          \State Introduce new concept $Z$
          \State \Return \{ $Q \sqcap R \sqsubseteq Z$, $Z \sqsubseteq Q$, $Z \sqsubseteq R$, $Z \sqsubseteq sp$ \}
 \EndIf
 \EndIf
	\end{algorithmic}
	\label{alg-normalization}
\end{algorithm}

\subsection{Weakening and completing}

\begin{figure*}[!ht] 
\begin{center}

\includegraphics[scale=0.4]{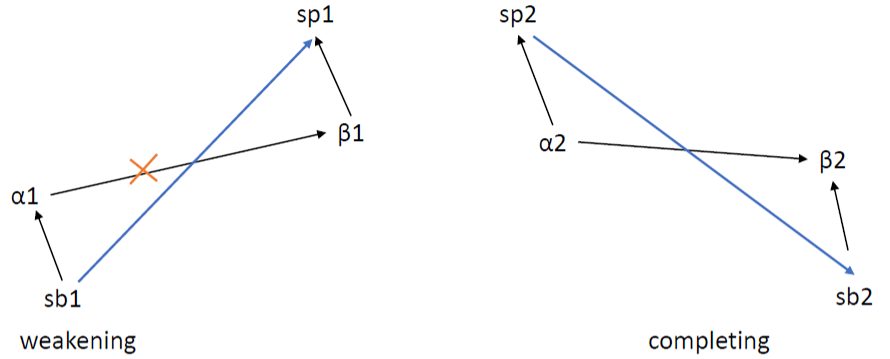}

\caption{Examples. Weakening: unwanted axiom $\alpha1$ $\sqsubseteq$ $\beta1$ is replaced by correct axiom $sb1$ $\sqsubseteq$ $sp1$; assumed that $\alpha1$ $\sqsubseteq$ $sp1$ is not correct; formerly derivable correct axiom $sb1$ $\sqsubseteq$ $sp1$ still entailed by repaired ontology. 
Completion: wanted axiom $\alpha2$ $\sqsubseteq$ $\beta2$ is replaced by correct axiom $sp2$ $\sqsubseteq$ $sb2$; $\alpha2$ $\sqsubseteq$ $\beta2$ is still derivable and additional correct axiom $sp2$ $\sqsubseteq$ $sb2$ in the repaired ontology.} 
\label{fig-weakening-completing}    
\end{center}
\end{figure*}

Given an axiom, \textit{weakening} aims to find other axioms that are weaker than the given axiom, i.e., the given axiom logically implies the other axioms. For an axiom $\alpha$ $\sqsubseteq$ $\beta$, this is often done by replacing $\alpha$ by a more specific concept or replacing $\beta$ by a more general concept.
For the repairing this means that a wrong axiom $\alpha$ $\sqsubseteq$ $\beta$ can be replaced by a correct weaker axiom, thereby mitigating the effect of removing the wrong axiom (Figure \ref{fig-weakening-completing}). Algorithm \ref{alg-weakened-axiom-set} presents a tractable weakening algorithm for normalized $\el$ TBoxes. For a given axiom $\alpha$ $\sqsubseteq$ $\beta$, it finds correct axioms $sb$ $\sqsubseteq$ $sp$ such that $sb$ is a sub-concept in SCC(${\cal T})$ of $\alpha$ and $sp$ is a super-concept in SCC(${\cal T}$) of $\beta$. Further, there should not be another correct axiom under these conditions that would add more correct knowledge to the ontology than $sb$ $\sqsubseteq$ $sp$. As we work with normalized $\el$ TBoxes, the new axioms are normalized. The existence of such weaker axioms is not guaranteed.

\textit{Completing} aims to find correct axioms that are not derivable from the ontology yet and that would make a given axiom derivable. It was introduced to aid domain experts when adding axioms to the ontology to find additional knowledge to add. While weakening is usually performed on unwanted axioms, completing is usually performed on wanted axioms.
Algorithm \ref{alg-completed-axiom-set} presents a tractable completion algorithm for normalized $\el$ TBoxes. For a given axiom $\alpha$ $\sqsubseteq$ $\beta$, it finds correct axioms $sp$ $\sqsubseteq$ $sb$ such that $sp$ is a super-concept in SCC(${\cal T})$ of $\alpha$ and $sb$ is a sub-concept in SCC(${\cal T}$) of $\beta$ (Figure \ref{fig-weakening-completing}). This means that if $sp$ $\sqsubseteq$ $sb$ is added to ${\cal T}$, then $\alpha$ $\sqsubseteq$ $\beta$ would be derivable. Further, there should not be another correct axiom under these conditions that would add more correct knowledge to the ontology than $sp$ $\sqsubseteq$ $sb$. Similarly as for weakening, the new axioms are normalized. The completed axiom set is guaranteed to be not empty for a correct axiom $\alpha$ $\sqsubseteq$ $\beta$. It will contain $\alpha$ $\sqsubseteq$ $\beta$ or other axioms that lead to the derivation of $\alpha$ $\sqsubseteq$ $\beta$.

Note that weakening and completing are dual operations where the former finds weaker axioms and the latter stronger axioms. This is reflected in the mirroring of the sub- and super-concepts of $\alpha$ and $\beta$ in Algorithms 	\ref{alg-weakened-axiom-set} and \ref{alg-completed-axiom-set}.

\begin{algorithm}

	\caption{Weakened axiom set} 
	\hspace*{\algorithmicindent}\textbf{Input}: TBox ${\cal T}$,  Oracle Or, unwanted axiom $\alpha$ $\sqsubseteq$ $\beta$ \\
   \hspace*{\algorithmicindent}\textbf{Output}: The weakened axiom set of $\alpha$ $\sqsubseteq$ $\beta$
	\begin{algorithmic}[1]
   \State $wt_{\alpha \sqsubseteq \beta}\leftarrow$\{$sb$ $\sqsubseteq$ $sp$ $\mid$ 
   $sb$ $\in$ $sub(\alpha,{\cal T})$ 
   $\wedge$ $sp$ $\in$ $sup(\beta,{\cal T})$
   $\wedge$ Or($sb \sqsubseteq sp$) = True 
   $\wedge$ $\neg$$\exists$ $sb'$ $\in$ $sub(\alpha,{\cal T})$, $sp'$ $\in$ $sup(\beta,{\cal T})$: (Or($sb' \sqsubseteq sp'$)  = True $\wedge$ (($sb$ $\sqsubseteq$ $sb'$ $\wedge$ $sp'$ $\sqsubset$ $sp$) $\vee$ ($sb$ $\sqsubset$ $sb'$ $\wedge$ $sp'$ $\sqsubseteq$ $sp$)))\}
   \State $w_{\alpha \sqsubseteq \beta}\leftarrow$ $\emptyset$
   \ForEach {$sb$ $\sqsubseteq$ $sp$ $\in$ $wt_{\alpha \sqsubseteq \beta}$}
   \State $w_{\alpha \sqsubseteq \beta} \leftarrow$ $w_{\alpha \sqsubseteq \beta}\cup$ Normalize($sb$ $\sqsubseteq$ $sp$)
    \EndFor
    \State \Return $w_{\alpha \sqsubseteq \beta}$
	\end{algorithmic}
\label{alg-weakened-axiom-set}
\end{algorithm}

\begin{algorithm}
	\caption{Completed axiom set} 
	\hspace*{\algorithmicindent}\textbf{Input}: TBox ${\cal T}$,  Oracle Or, a wanted axiom $\alpha$ $\sqsubseteq$ $\beta$ \\
   \hspace*{\algorithmicindent}\textbf{Output}: The completed axiom set of $\alpha$ $\sqsubseteq$ $\beta$
	\begin{algorithmic}[1]
   \State $ct_{\alpha \sqsubseteq \beta}\leftarrow$\{$sp$ $\sqsubseteq$ $sb$ $\mid$ $sp$ $\in$ $sup(\alpha,{\cal T})$
   $\wedge$ $sb$ $\in$ $sub(\beta,{\cal T})$
   $\wedge$ Or($sp \sqsubseteq sb$) = True 
   $\wedge$ $\neg$$\exists$ $sp'$ $\in$ $sup(\alpha)$, $sb'$ $\in$ $sub(\beta)$: (Or($sp' \sqsubseteq sb'$)
  = True $\wedge$ ($sp$ $\sqsubseteq$ $sp'$ $\wedge$ $sb'$ $\sqsubset$ $sb$) $\vee$ ($sp$ $\sqsubset$ $sp'$ $\wedge$ $sb'$ $\sqsubseteq$ $sb$)\}
    \State $c_{\alpha \sqsubseteq \beta}\leftarrow$ $\{ \alpha \sqsubseteq \beta \}$
   \ForEach {$sb$ $\sqsubseteq$ $sp$ $\in$ $ct_{\alpha \sqsubseteq \beta}$}
   \State $c_{\alpha \sqsubseteq \beta} \leftarrow$ $c_{\alpha \sqsubseteq \beta}\cup$ Normalize($sb$ $\sqsubseteq$ $sp$)
    \EndFor
    \State \Return $c_{\alpha \sqsubseteq \beta}$
	\end{algorithmic}
	\label{alg-completed-axiom-set}
\end{algorithm}

\subsection{Combinations}
\label{sec-combinations}

Given a set of wrong axioms, there are different ways to repair the ontology using the removing, weakening and completing operations. There are choices to be made regarding the use of wrong axioms in the weakening and completing steps, regarding removing, weakening and completing all axioms at once or one at a time and in the latter case regarding the order the axioms are processed, as well as regarding when to update the ontology. Each of these choices may have an influence on the completeness and incorrectness of the repaired ontology. In general, using as much (possibly wrong) information as possible may lead to more complete ontologies, but also requires a larger validation effort.

\subsubsection{Algorithms}

We show different algorithms (Algorithms \ref{alg-RW1-WAS} to \ref{alg-R-WCCAS1}) 
that use different ways of combining removing, weakening and completing.
 (The algorithms that are used for the discussion are shown here. All algorithms we used in the experiments are shown in appendix A.)
In Algorithms \ref{alg-RW1-WAS} to \ref{alg-R-WWAS1} we only remove and weaken, while completing is added in Algorithms
\ref{alg-RW1-C1-CAS} to \ref{alg-R-WCCAS1}. 
The operations that are used for the algorithms are shown in Table \ref{tab:algorithm-overview}.
We note that all proposed algorithms are tractable and find repairs as defined in Definition \ref{def-repair-ontology}.

\begin{algorithm}[ht!]
\renewcommand{\thealgorithm}{C1}
	\caption{Weaken one at a time, add weakened axiom sets and remove all wrong at end} 
     \hspace*{\algorithmicindent}\textbf{Input}: TBox ${\cal T}$,  Oracle Or, set of unwanted axioms $W$ \\
   \hspace*{\algorithmicindent}\textbf{Output}: A repaired TBox
	\begin{algorithmic}[1]
	\ForEach {$\alpha$ $\sqsubseteq$ $\beta$ $\in$ $W$}
	\State ${\cal T}_r$ $\leftarrow$ Remove-axioms(${\cal T}$, $\{\alpha \sqsubseteq\beta\}$)
   \State $w_{\alpha \sqsubseteq \beta}\leftarrow$ weakened-axiom-set($\alpha \sqsubseteq \beta$, ${\cal T}_r, Or)$
   \EndFor
  \State ${\cal T}_r$ $\leftarrow$  Add-axioms(${\cal T}$,$\bigcup_{\alpha \sqsubseteq\beta}$ $w_{\alpha \sqsubseteq \beta}$)
   \State \Return Remove-axioms(${\cal T}_r$,$W$)
	\end{algorithmic}
	\label{alg-RW1-WAS}
\end{algorithm}

\begin{algorithm}[h!]
\renewcommand{\thealgorithm}{C2}
	\caption{Remove/weaken/add weakened axiom sets one at a time}
     \hspace*{\algorithmicindent}\textbf{Input}: TBox ${\cal T}$,  Oracle Or, set of unwanted axioms $W$ \\
   \hspace*{\algorithmicindent}\textbf{Output}: A repaired TBox
	\begin{algorithmic}[1]
	\State ${\cal T}_r$ $\leftarrow$ ${\cal T}$
	\ForEach {$\alpha$ $\sqsubseteq$ $\beta$ $\in$ $W$}
	\State ${\cal T}_r$ $\leftarrow$ Remove-axioms(${\cal T}_r$, $\{\alpha \sqsubseteq\beta\}$)
   \State $w_{\alpha \sqsubseteq \beta}\leftarrow$ weakened-axiom-set($\alpha \sqsubseteq \beta$, ${\cal T}_r, Or)$
   \State ${\cal T}_r$ $\leftarrow$ Add-axioms(${\cal T}_r$,$w_{\alpha \sqsubseteq \beta}$)
   \EndFor
   \State \Return ${\cal T}_r$
	\end{algorithmic}
	\label{alg-RWWAS1}
	\end{algorithm}

\begin{algorithm}[h!]
\renewcommand{\thealgorithm}{C3}
	\caption{Remove all wrong, weaken all and add weakened axiom sets at end}
     \hspace*{\algorithmicindent}\textbf{Input}: TBox ${\cal T}$,  Oracle Or, set of unwanted axioms $W$ \\
   \hspace*{\algorithmicindent}\textbf{Output}: A repaired TBox
	\begin{algorithmic}[1]
	\State ${\cal T}_r$ $\leftarrow$ Remove-axioms(${\cal T}$, $W$)
	\ForEach {$\alpha$ $\sqsubseteq$ $\beta$ $\in$ $W$}
   \State $w_{\alpha \sqsubseteq \beta}\leftarrow$ weakened-axiom-set($\alpha \sqsubseteq \beta$, ${\cal T}_r, Or)$
   \EndFor
   \State \Return Add-axioms(${\cal T}_r$,$\bigcup_{\alpha \sqsubseteq\beta}$ $w_{\alpha \sqsubseteq \beta}$)
	\end{algorithmic}
	\label{alg-R-W1-WAS}
	\end{algorithm}

\begin{algorithm}[ht!]
\renewcommand{\thealgorithm}{C4}
	\caption{Remove all wrong, weaken/add weakened axiom sets one at a time}
     \hspace*{\algorithmicindent}\textbf{Input}: TBox ${\cal T}$,  Oracle Or, set of unwanted axioms $W$ \\
   \hspace*{\algorithmicindent}\textbf{Output}: A repaired TBox
	\begin{algorithmic}[1]
	\State ${\cal T}_r$ $\leftarrow$ Remove-axioms(${\cal T}$, $W$)
	\ForEach {$\alpha$ $\sqsubseteq$ $\beta$ $\in$ $W$}
   \State $w_{\alpha \sqsubseteq \beta}\leftarrow$ weakened-axiom-set($\alpha \sqsubseteq \beta$, ${\cal T}_r, Or)$
   \State ${\cal T}_r$ $\leftarrow$ Add-axioms(${\cal T}_r$,$w_{\alpha \sqsubseteq \beta}$)
   \EndFor
   \State \Return ${\cal T}_r$
	\end{algorithmic}
	\label{alg-R-WWAS1}
	\end{algorithm}

\begin{algorithm}[h!]
\renewcommand{\thealgorithm}{C9}
	\caption{Weaken one at a time, remove all wrong, complete one at a time, then add completed axiom sets at end}
     \hspace*{\algorithmicindent}\textbf{Input}: TBox ${\cal T}$,  Oracle Or, set of unwanted axioms $W$ \\
   \hspace*{\algorithmicindent}\textbf{Output}: A repaired TBox
	\begin{algorithmic}[1]
	\ForEach {$\alpha$ $\sqsubseteq$ $\beta$ $\in$ $W$}
	\State ${\cal T}_r$ $\leftarrow$ Remove-axioms(${\cal T}$, $\{\alpha \sqsubseteq\beta\}$)
   \State $w_{\alpha \sqsubseteq \beta}\leftarrow$ weakened-axiom-set($\alpha \sqsubseteq \beta$, ${\cal T}_r, Or)$
   \EndFor
    \State ${\cal T}_r$ $\leftarrow$ Remove-axioms(${\cal T}_r$,$W$)
   	\ForEach {$\alpha$ $\sqsubseteq$ $\beta$ $\in$ $W$}
   	\State $c_{\alpha \sqsubseteq \beta}\leftarrow$ $\emptyset$
	\ForEach {$sb$ $\sqsubseteq$ $sp$ $\in$ $w_{\alpha \sqsubseteq \beta}$}
   \State $c_{sb \sqsubseteq sp}\leftarrow$ completed-axiom-set($sb \sqsubseteq sp$, ${\cal T}_r, Or)$
   \State $c_{\alpha \sqsubseteq \beta}\leftarrow$ $c_{\alpha \sqsubseteq \beta}$ $\cup$ $c_{sb \sqsubseteq sp}$
   \EndFor
  \EndFor
  \State ${\cal T}_r$ $\leftarrow$  Add-axioms(${\cal T}_r$,$\bigcup_{\alpha \sqsubseteq\beta}$ $c_{\alpha \sqsubseteq \beta}$)
   \State \Return ${\cal T}_r$
	\end{algorithmic}
	\label{algorithm9_V1}
\end{algorithm}

\begin{algorithm}[h!]
\renewcommand{\thealgorithm}{C10}
	\caption{Weaken one at a time, remove all wrong, complete/add completed axiom sets one at a time}
     \hspace*{\algorithmicindent}\textbf{Input}: TBox ${\cal T}$,  Oracle Or, set of unwanted axioms $W$ \\
   \hspace*{\algorithmicindent}\textbf{Output}: A repaired TBox
	\begin{algorithmic}[1]
	\ForEach {$\alpha$ $\sqsubseteq$ $\beta$ $\in$ $W$}
	\State ${\cal T}_r$ $\leftarrow$ Remove-axioms(${\cal T}$, $\{\alpha \sqsubseteq\beta\}$)
   \State $w_{\alpha \sqsubseteq \beta}\leftarrow$ weakened-axiom-set($\alpha \sqsubseteq \beta$, ${\cal T}_r, Or)$
   \EndFor
   \State ${\cal T}_r$ $\leftarrow$ Remove-axioms(${\cal T}$, $W$)
   	\ForEach {$\alpha$ $\sqsubseteq$ $\beta$ $\in$ $W$}
	\ForEach {$sb$ $\sqsubseteq$ $sp$ $\in$ $w_{\alpha \sqsubseteq \beta}$}
   \State $c_{sb \sqsubseteq sp}\leftarrow$ completed-axiom-set($sb \sqsubseteq sp$, ${\cal T}_r, Or)$
   \State ${\cal T}_r$ $\leftarrow$  Add-axioms(${\cal T}_r$,$c_{sb \sqsubseteq sp}$)
   \EndFor
  \EndFor
   \State \Return ${\cal T}_r$
	\end{algorithmic}
	\label{alg-RW1-CCAS1}
\end{algorithm} 

\clearpage

\subsubsection{Comparing strategies}

To show the trade-off between the choices regarding completeness and validation effort between the different algorithms, we define operators in Table  \ref{tab:operation-overview} that can be used as building blocks in the design of algorithms. The operations represent choices regarding the use of wrong axioms by removing them (R) and adding them back (AB), regarding weakening (W) and completing (C) one at a time or all at once. Furthermore, update (U) is always used in combination with weakening or completing, and relates to when changes to the ontology are performed. For instance, when weakening an axiom, the weakened axioms could be added immediately to the ontology (and thus influence the ontology before weakening other axioms) or can be added after having weakened all axioms (and thus weakening one axiom does not influence weakening the next axiom).\footnote{In the algorithms this is represented by the use of $\mathcal{T}$ or $\mathcal{T}_r$ as TBox when computing weakened axioms or completed axioms sets.}
The combination algorithms can be defined by which of these building blocks are used and in which order. For instance,  Algorithm \ref{algorithm9_V1} uses weaken one at a time, remove all wrong, complete one at a time, then add completed axiom sets at the end, while Algorithm \ref{alg-RW1-CCAS1} uses weaken one at a time, remove all wrong, add completed axiom sets one at a time. 

\begin{figure*}[!ht] 
\begin{center}
\subfigure[Removing]{\includegraphics[width=0.18
\textwidth]{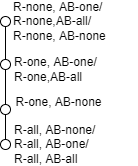}}\label{a}
\subfigure[Weakening]{\includegraphics[width=0.18\textwidth]{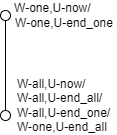}} \label{b}
\subfigure[Completing]{\includegraphics[width=0.18\textwidth]{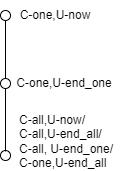}}\label{c} 
\caption{Hasse diagrams. (a) shows ways to remove and add back wrong axioms; (b) weakening and when to update the ontology; (c) completing and when to update the ontology. Combinations of operations higher up in the lattices lead to more validation work and more complete ontologies.} 
\label{fig-lattice}    
\end{center}
\end{figure*}

 The operations have different effects on the completeness of the final ontology and validation effort. This is represented in the Hasse diagrams in Figure \ref{fig-lattice} where the partial order represents more or equally complete final ontologies.
For instance, Figure \ref{fig-lattice}b  shows that weakening one axiom at a time and immediately updating the TBox (W\_one,U\_now) leads to a more complete ontology (and more validation effort) than the other choices.
Figure \ref{fig-lattice}c, shows that ontologies repaired by algorithms using one axiom at a time completing and immediate updates (C-one,U-now) are more complete than ontologies repaired using one axiom at a time completing and updating the ontology after each weakened axiom set for a wrong axiom (C-one,U-end\_one). These ontologies are in turn more complete than for the other choices.
Similar observations regarding removing wrong axioms are in Figure  \ref{fig-lattice}a. How the Hasse diagrams are derived, is explained in appendix C.

Using these Hasse diagrams, we can then compare algorithms. If the sequence of operators for one algorithm can be transformed to the sequence of operators of a second algorithm, by replacing some operators of the first algorithm using operators higher up in the lattices in  Figure \ref{fig-lattice}, then the ontologies repaired using the second algorithm will be more (or equally) complete than the ontologies repaired using the first algorithm. For instance, the sequence of Algorithm \ref{algorithm9_V1} 
can be rewritten into the sequence of Algorithm \ref{alg-RW1-CCAS1} 
by replacing the completion operator to a higher-level completion operator. Thus, repairing an ontology using Algorithm 
\ref{alg-RW1-CCAS1} 
will lead to a more (or equally)  complete ontology than repairing using Algorithm \ref{algorithm9_V1}, but also requires more validation work.

\begin{table*}[ht!]
\begin{center}
\caption{\label{tab:operation-overview}Different operations for removing, weakening and completing.}
\begin{tabular}{|l|l|}
\hline
Operations & Description \\
\hline
R-all & Remove all the wrong axioms at once.\\
R-one & Remove the wrong axioms one at a time.\\
R-none & Remove nothing.\\

\hline
\hline
W-all & Weaken all wrong axioms at once.\\
W-one & Weaken the wrong axioms one at a time\\
\hline
\hline
C-all & Complete all weakened axioms at once.\\
C-one & Complete the weakened axioms one at a time.\\
\hline
\hline
AB-one & Add one wrong axiom back.\\ 
AB-all & Add all wrong axioms back.\\
AB-none & Add nothing back.\\
\hline
\hline
U-now & Update the changes immediately. \\
U-end\_one & Update the changes after the iteration of each wrong axiom. \\
U-end\_all & Update the changes after iterations of all wrong axioms. \\
\hline
\end{tabular}
\end{center}

\end{table*}

\section{Experiments} 
\label{sec-experiments}

In order to compare the use of the different combinations of strategies, 
as defined in section \ref{sec-combinations}, 
we run experiments on several ontologies.\footnote{Mini-GALEN, an example inspired by the GALEN (from \url{https://bioportal.bioontology.org/}) ontology;
PACO, NCI, OFSMR available at \url{https://bioportal.bioontology.org/};
EKAW from the conference track of  \url{http://oaei.ontologymatching.org/}; Pizza ontology available at \url{https://github.com/owlcs/pizza-ontology}.}
An overview of the numbers of concepts, roles and axioms in these ontologies is given in Table \ref{tab:ontology}.
We have used the parts of these ontologies that are expressible in $\el$ in the sense that we removed the parts of axioms that used constructors not in $\el$. We introduced  3 to 6 wrong axioms in each ontology by replacing existing axioms with axioms where the left-hand or right-hand side concepts of the existing axioms were changed. 
For subsumption checking in the algorithms we used 
HermiT\footnote{http://www.hermit-reasoner.com/}. 
For the full results of the experiments we refer to appendix B while we discuss and exemplify
interesting observations based on  
for Mini-GALEN (Figure \ref{fig-Mini-GALEN} and 
visualized in Figure 
\ref{fig-Mini-GALEN-viz}).
Table \ref{tab:comparison1} shows results for Algorithms \ref{alg-RW1-WAS}/\ref{alg-RWWAS1} vs \ref{alg-R-W1-WAS}/\ref{alg-R-WWAS1} 
regarding the number 
of sub-concepts of $\alpha$ and super-concepts of $\beta$ for each wrong axiom $\alpha \sqsubseteq \beta$ when choosing to remove one wrong axiom at a time or all at once. In the table for each algorithm there is one sub and one sup set for each of the wrong axioms 
(e.g., for \ref{alg-RW1-WAS} for the first wrong axiom there are 3 concepts in the sup set and 2 in the sub set, resulting in 6 candidate weakened axioms). 
Further, the weakened axioms are shown.
Table \ref{tab:comparison-order2} shows the sizes of the sub and sup  sets and the axioms to add using different orders of computing weakened axioms sets and adding them as soon as they are found for Algorithm \ref{alg-R-WWAS1}.
In Table \ref{tab:example2 } we show the sizes of the sub and sup  sets for the completing step as well as the completed axioms for Algorithms \ref{algorithm9_V1} and \ref{alg-RW1-CCAS1}.

\begin{table}[t]
\begin{center}
\caption{\label{tab:ontology}Ontologies}
\begin{tabular}{|l|r|r|r|r|r|r|}
\hline
&Mini-&Pizza&EKAW&OFSMR&PACO&NCI\\
&GALEN&&&&&\\\hline
 Concepts&9&74&100&159&224&3304\\\hline
Roles&1&33&8&2&23&1\\\hline Axioms&20&341&801&1517&1153&30364\\\hline
\end{tabular}
\end{center}

\end{table}

\begin{figure*}[!ht]
\begin{center}
\begin{tabular}{|l|} \hline
$N_C$ = \{ GPr (GranulomaProcess), CVD (CardioVascularDisease),\\ PPh (PathologicalPhenomenon),  
F (Fracture), E (Endocarditis), C (Carditis), \\IPr (InflammationProcess), PPr (PathologicalProcess),
NPr (NonNormalProcess)\} \\
$N_R$ = \{ hAPr (hasAssociatedProcess) \} \\

${\cal T}$ = \{ 
CVD $\sqsubseteq$ PPh, 
F $\sqsubseteq$ PPh, 
$\exists$hAPr.PPr $\sqsubseteq$ PPh, 
E $\sqsubseteq$ C, 
E $\sqsubseteq$ $\exists$hAPr.IPr, 
GPr $\sqsubseteq$ NPr, \\
PPr $\sqsubseteq$ IPr,
IPr $\sqsubseteq$ GPr, 
E $\sqsubseteq$ PPr
\} \\

$W$ = \{
E $\sqsubseteq$ PPr,
PPr $\sqsubseteq$ IPr,
IPr $\sqsubseteq$ GPr
\} \\


$Or$ returns $true$ for: 
GPr  $\sqsubseteq$ IPr, 
GPr  $\sqsubseteq$ PPr, 
GPr  $\sqsubseteq$ NPr, 
IPr $\sqsubseteq$ PPr, 
IPr $\sqsubseteq$ NPr,\\
PPr $\sqsubseteq$ NPr,
CVD $\sqsubseteq$ PPh,
F $\sqsubseteq$ PPh, 
E $\sqsubseteq$ PPh, 
E $\sqsubseteq$ C, 
E $\sqsubseteq$ CVD, 
C $\sqsubseteq$ PPh, 
C $\sqsubseteq$ CVD, \\
$\exists$hAPr.PPr $\sqsubseteq$ PPh, 
$\exists$hAPr.IPr $\sqsubseteq$ PPh, 
E $\sqsubseteq$ $\exists$hAPr.IPr,
E $\sqsubseteq$ $\exists$hAPr.PPh.\\
Note that for an oracle that does not make mistakes, 
if  $Or$(P $\sqsubseteq$ Q) = true,\\ then also $Or$($\exists$r.P $\sqsubseteq$ $\exists$r.Q)=true
and  $Or$(P $\sqcap$ O $\sqsubseteq$ Q)=true.\\
For other axioms P $\sqsubseteq$ Q with P, Q $\in$ $N_C$, $Or$(P $\sqsubseteq$ Q) = false.\\
\hline
\end{tabular}
\end{center}
\caption{Mini-GALEN. (Visualized in Figure \ref{fig-Mini-GALEN-viz}.)}
\label{fig-Mini-GALEN}
\end{figure*}

\begin{figure*}[!ht] 
\begin{center}
\includegraphics[width=0.6\textwidth]{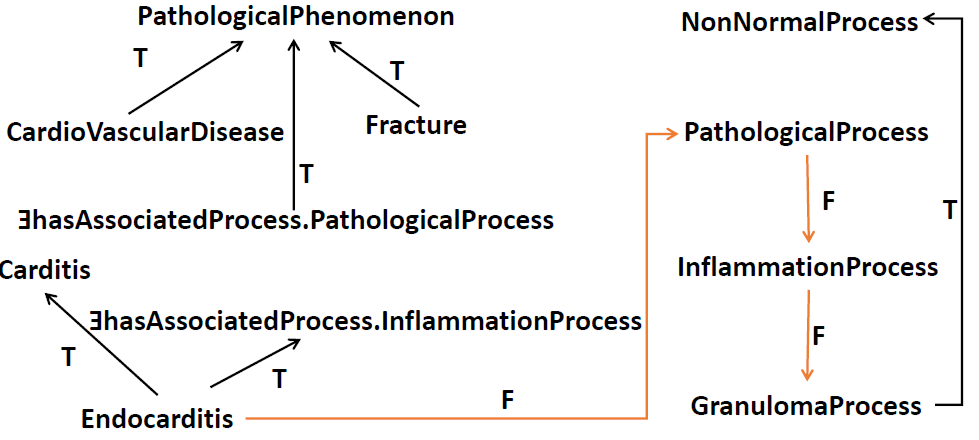}

\caption{Visualization of the Mini-GALEN ontology in Figure \ref{fig-Mini-GALEN}. The axioms in the TBox are represented with black arrows except for the wrong axioms which are represented in red. The oracle's knowledge about the axioms in the ontology is marked with T (true) or F (false) at the arrows.} 
\label{fig-Mini-GALEN-viz}
\end{center}
\end{figure*}

\begin{table}[t]
\begin{center}
\caption{\label{tab:comparison1}Weakening for Mini-GALEN using Algorithms \ref{alg-RW1-WAS}-\ref{alg-R-WWAS1}. Three wrong axioms give 3 sup/sub-sets per algorithm.}
\begin{tabular}{|p{1.4cm}|p{1.6cm}|p{1.6cm}|p{1.6cm}|p{1.6cm}|}
\hline
 &C1&C2&C3&C4\\\hline
 Sup($\beta$,${\cal T}$)&3  2  4&3  2  2&1  2  1&1  2  1\\
 Sub($\alpha$,${\cal T}$)&2 3 1&2 1 1&1 1 1&1 1 1\\\hline
 Weakened
 & PPr $\sqsubseteq$ NPr IPr $\sqsubseteq$ NPr  
 & PPr $\sqsubseteq$ NPr IPr $\sqsubseteq$ NPr
 & IPr $\sqsubseteq$ NPr 
 & IPr $\sqsubseteq$ NPr\\
 \hline
\end{tabular}
\end{center}

\end{table}

\begin{table*}[!h]
\begin{center}
\caption{\label{tab:comparison-order2} Adding weakened axioms in different order for Mini-GALEN by Algorithm \ref{alg-R-WWAS1}. Wrong axioms:  
\textcircled{1}PPr$\sqsubseteq$IPr, \textcircled{2}IPr$\sqsubseteq$GPr, \textcircled{3}E$\sqsubseteq$PPr.}
\begin{tabular}{|p{1.35cm}|p{1.6cm}|p{1.6cm}|p{1.6cm}|p{1.6cm}|p{1.6cm}|p{1.6cm}|}
\hline
 Wrong &\textcircled{1}$\rightarrow$\textcircled{2}$\rightarrow$\textcircled{3}&\textcircled{1}$\rightarrow$\textcircled{3}$\rightarrow$\textcircled{2}&\textcircled{2}$\rightarrow$\textcircled{1}$\rightarrow$\textcircled{3} &\textcircled{2}$\rightarrow$\textcircled{3}$\rightarrow$\textcircled{1} &\textcircled{3}$\rightarrow$\textcircled{1}$\rightarrow$\textcircled{2} &\textcircled{3}$\rightarrow$\textcircled{2}$\rightarrow$\textcircled{1} \\\hline
 Sup($\beta$,${\cal T}$)&1  2  1&1  2  1&2  2 2&2 2 1&1  2  1&3 2 1  \\
 Sub($\alpha$,${\cal T}$)&1 1 1&1 1 1&1 1 1&1 1 1&1 1 1&1 1 1\\\hline
 Weakened
 & IPr $\sqsubseteq$ NPr
 & IPr $\sqsubseteq$ NPr
 & IPr $\sqsubseteq$ NPr PPr $\sqsubseteq$ NPr
 & IPr $\sqsubseteq$ NPr PPr $\sqsubseteq$ NPr
 & IPr $\sqsubseteq$ NPr
 & IPr $\sqsubseteq$ NPr PPr $\sqsubseteq$ NPr
\\\hline
\end{tabular}
\end{center}

\end{table*}

\begin{table}[!h]
\begin{center}
\caption{\label{tab:example2 }Completing Mini-GALEN using Algorithms C9-C10.}
\begin{tabular}{|l|r|r|r|r|}
\hline
 &C9&C10\\\hline
Sup($\alpha$,${\cal T}$)&1 1&1 1\\
Sub($\beta$,${\cal T}$)&2 2&2 3\\\hline
Completed
&PPr$\sqsubseteq$NPr, IPr$\sqsubseteq$NPr
&PPr$\sqsubseteq$NPr, IPr$\sqsubseteq$PPr
\\\hline
\end{tabular}
\end{center}

\end{table}

\section{Discussion} 
\label{sec-discussion}

{\bf Choosing an algorithm.}
The most preferred repair for an ontology with wrong axioms would lead to a more complete and less incorrect ontology than the original ontology. In general, however, this cannot be guaranteed unless we use a brute-force method that checks all axioms in an ontology. Although some optimizations are possible, this is in general not feasible. On the positive side, removing axioms will not introduce more incorrect knowledge and adding axioms will not remove correct knowledge. Unfortunately, removing wrong axioms may make the ontology less complete. For instance, when removing $W$ from Mini-GALEN, the correct axiom PPr $\sqsubseteq$ NPr cannot be derived anymore.
The weakening and completing alleviate this problem, but do not solve it completely. Adding correct axioms may make the ontology more incorrect in the case where some defects in the ontology were not yet detected or repaired and these lead to the derivation of new defects.

There is also a trade-off between using as much, but possibly wrong, knowledge as possible in the ontology and removing as much wrong knowledge as possible, when computing weakened and completed axiom sets (Figure \ref{fig-lattice}a). In the former case, more axioms (including more wrong axioms) are generated and need to be validated than in the latter case, but the final ontology in the former case may be more complete than in the latter case. 
For instance, Table \ref{tab:comparison1} shows that sizes of the sup and sub sets for removing one axiom at a time are larger than or equal to the sizes of the sets for removing all at once (Algorithms \ref{alg-RW1-WAS}/\ref{alg-RWWAS1} vs \ref{alg-R-W1-WAS}/\ref{alg-R-WWAS1}).
When removing one at a time, the other wrong axioms can lead to more sub- and super-concepts and thus larger weakened axiom sets. This entails a higher validation effort by the domain expert, but it also leads to a more complete ontology as the axiom \textit{PPr $\sqsubseteq$ NPr} is not (always, see below) found by the approaches that remove all at once.

Another choice is to add new correct axioms as soon as they are found or wait until the end (Figure \ref{fig-lattice}b). In the former case they may be used to find additional information, but the end result may depend on the order that the axioms are handled. Also the order in which the axioms are processed, has an influence on the result as seen from Table \ref{tab:comparison-order2} for Algorithm \ref{alg-R-WWAS1}.
Similar observations can be made when completing is added to the removing and weakening (Figure \ref{fig-lattice}c).
When the completed axioms are added one at a time during each iteration, the sizes of the sub and sup sets for each weakened axiom will be larger than or equal to the sizes of the sets generated when adding them all at the end. In Table \ref{tab:example2 } we show this for Algorithms \ref{algorithm9_V1} and \ref{alg-RW1-CCAS1} 
that differ from each other in this aspect.
Also here it entails a higher validation effort by the domain expert when adding one at a time, but it also leads to a more complete ontology.

 In general, there is a trade-off between validation effort and completeness and thus the choice of algorithm depends on the priority between these. Earlier work in the field usually discussed one algorithm and did not show there were different options. In this work we show this trade-off. Further, by providing the Hasse diagrams we help deciding which features to use. Using features higher up in the diagrams means more validation work and more complete ontologies.

{\bf Domain expert validation in practice.} 
The introduction of new concepts 
may make it hard for a domain expert to validate the axioms, as mentioned in \cite{BKNP18}. In our implemented systems we, therefore, alleviate this problem by using a naming convention that reflects the logical description of the new concepts as they would be in a non-normalized TBox. For instance, we use names such as 'S-SOME-Q' and 'Q-AND-R'. This convention also allows the nesting of operators. In future work we will investigate the technique of 'forgetting' (e.g., \cite{Wang2009ConceptAR}) to further alleviate this problem. 

In \cite{BKNP18} an ontology consists of a static part considered to be correct and a refutable part. If we would follow this approach, then in our setting the set of wrong axioms can only be from the refutable part and axioms from the static part should never be removed. Axioms from the static part also do not need to be validated. Adding correct axioms should then grow the static part.
We note that, in practise, it is not so clear how to divide an ontology in a static and a refutable part as, as mentioned before, according to our experience in assisting the development of ontologies in different domains, domain experts make mistakes even in the parts they think are correct.

\section{Implemented systems} 
\label{sec-systems}


We have implemented two systems. As Prot\'{e}g\'{e} is a well-known ontology development tool, we implemented a plugin for repairing based on Algorithm C9. 
Using this algorithm the user can repair all wrong axioms at once. However, by iteratively invoking this plugin the user can also repair the wrong axioms one at a time.
Further, we extended the $\mathcal{EL}$ version of the RepOSE system \cite{WDL14,Lambrix2015CompletingTI}. We allow the user to choose different combinations, thereby giving a choice in the trade-off between validation work and completeness.
In the system candidate weakened and completing axioms are shown in lists and also visualized using two sets of concepts.
The axioms $\alpha$ $\sqsubseteq$ $\beta$ to be validated are the ones that can be constructed by choosing $\alpha$ from the first set and $\beta$ from the second set. By showing them together, context of the solutions in the form of sub- and super-concepts is available. The domain expert can choose to validate such axioms (and choose the best ones) by clicking in the different panes representing the sets of concepts. 
Furthermore, to reduce the amounts of concepts in the completed axiom sets we only show combinations that would not introduce equivalence between concepts in the ontology. This means that in the implemented version of Algorithm \ref{alg-completed-axiom-set}, $sp$ should belong to $sup(\alpha,{\cal T})$ $\setminus$ $sup(\beta,{\cal T})$ ($sup(\alpha,{\cal T})$ in Algorithm \ref{alg-completed-axiom-set}) and $sb$ to  $sub(\beta,{\cal T})$ $\setminus$ $sub(\alpha,{\cal T})$ ($sub(\beta,{\cal T})$ in Algorithm \ref{alg-completed-axiom-set}). These new sets of super- and sub-concepts are called \textit{source} and \textit{target}.
In the supplemental material available at \url{https://www.ida.liu.se/~patla00/publications/EL-removing-weakening-completing/}, we provide the systems, user manuals and examples.

\section{Related work}
\label{sec-related-work}

We briefly discuss previous work on weakening and on completing. We are not aware of work that combines these. 
Regarding weakening, previous work looks at the combination of debugging and weakening. Justifications for wrong axioms and a hitting set are computed. Then, instead of removing, weakened axioms are computed. In our approach we assume that the axioms to remove are given (e.g., by having computed a hitting set) and that when removing them they cannot be derived anymore. When this assumption is not made then, as pointed out in \cite{BKNP18} (and ignored by older approaches) the weakening needs to be iterated to obtain a repair. We also note that none of the approaches explicitly state the use of a domain expert/oracle and they are purely logic-based. In practice, however, a domain expert/oracle is needed as otherwise axioms that are wrong in the domain of the ontology could be added. 
In contrast to our approach, the other approaches work on non-normalized TBoxes. This means that they may find better solutions for the weakening, but the search space for solutions also becomes infinite. In \cite{TCGPPK18} algorithms for weakening for $\el$ and ${\cal ALC}$ are given with tractable and exponential complexity, respectively. They are based on refinement operators that are applied on the concepts of GCIs. The approach is extended in \cite{CGKPRT20} for ${\cal SROIQ}$ TBoxes with an algorithm with almost-sure termination. Also in \cite{DQF14} an approach based on refinement operators is presented for ${\cal ALC}$. The nesting of operators is restricted based on the size of a concept. In \cite{BKNP18} the right-hand side of axioms is generalized, but the left-hand side is not specialized to obtain a well-founded weakening relation (i.e., there is no infinite chain of weakenings). Essentially, our use of $sup(P,{\cal T})$ and $sub(P,{\cal T})$ in the weakening is a similar approach. As we have restricted the $sup(P,{\cal T})$ and $sub(P,{\cal T})$ to contain only concepts in SCC(${\cal T}$), we only have a finite number of possible axioms.
In all these other approaches usually one-at-a-time removing (R-one, AB-none) and weakening (W-one, U-now) is used. From our Hasse diagrams we can see that this means the most complete ontologies and most validation work for weakening, but neither the most nor the least complete ontologies for removing. Further, the issue of the influence of the order is not addressed. 
In \cite{LSPV08} parts of axioms to remove are pinpointed and harmful and helpful changes are defined.

Regarding completing, the previous work with validation by a domain expert (e.g., \cite{WDL14} for the $\el$ family, \cite{LDI12} for ${\cal ALC}$) allowed only axioms of the form $P$ $\sqsubseteq$ $Q$ where $P$ and $Q$ are atomic concepts in the completed axioms set while Algorithm \ref{alg-completed-axiom-set} allows $P$ and $Q$ to be in SCC(${\cal T}$) (and then normalizes).
A non-interactive solution that is independent of the constructors of the description logic (e.g., tested with ontologies with expressivity up to ${\cal SHOIN(D)}$) is proposed in \cite{DWM17}. 
This approach introduces justification patterns that can be instantiated with existing concepts or new (called 'fresh') concepts. 

\section{Conclusion}
\label{sec-conclusion}

In this paper we proposed an interactive approach using weakening and completing to mitigate the negative effects of removing wrong axioms in $\el$ ontologies.  We
presented new algorithms and studied the combination of removing, weakening and completing. We show that the approach mitigates the negative effects, but also that there is a trade-off between completeness and the amount of domain expert validation. 

For future work there are interesting extensions. A first extension is to allow to remove other axioms than the axioms in $W$, thereby effectively integrating the full debugging (instead of just removing) with weakening and completing. While we have extended the solution setting from axioms with only atomic concepts to axioms with concepts in SCC(${\cal T}$) for completing, it is interesting to look at other solutions while still maintaining a practically feasible validation work for the domain expert. Further, it is interesting to investigate the problem for more expressive description logics.




\paragraph*{Acknowledgements:} 
We thank Olaf Hartig for discussions leading to the Hasse diagrams. This work is financially supported by the Swedish e-Science Research Centre (SeRC), and the Swedish Research Council (Vetenskapsr{\aa}det, dnr 2018-04147).

\newpage

\bibliography{ref-short}
\bibliographystyle{splncs04}


\newpage

\section*{Appendix A - Algorithms for Combination strategies}
In this appendix, we give more details about the different algorithms used in the experiments.
We show all our algorithms for combining different removing, weakening and completing strategies  including the ones that were presented in the paper earlier. A brief description of each algorithm is shown in Table \ref{tab:algorithm-overview}. 

\begin{algorithm}[ht!]
\renewcommand{\thealgorithm}{C1}
	\caption{Weaken one at a time, add weakened axiom sets and remove all wrong at end} 
     \hspace*{\algorithmicindent}\textbf{Input}: TBox ${\cal T}$,  Oracle Or, set of unwanted axioms $W$ \\
   \hspace*{\algorithmicindent}\textbf{Output}: A repaired TBox
	\begin{algorithmic}[1]
	\ForEach {$\alpha$ $\sqsubseteq$ $\beta$ $\in$ $W$}
	\State ${\cal T}_r$ $\leftarrow$ Remove-axioms(${\cal T}$, $\{\alpha \sqsubseteq\beta\}$)
   \State $w_{\alpha \sqsubseteq \beta}\leftarrow$ weakened-axiom-set($\alpha \sqsubseteq \beta$, ${\cal T}_r, Or)$
   \EndFor
  \State ${\cal T}_r$ $\leftarrow$  Add-axioms(${\cal T}$,$\bigcup_{\alpha \sqsubseteq\beta}$ $w_{\alpha \sqsubseteq \beta}$)
   \State \Return Remove-axioms(${\cal T}_r$,$W$)
	\end{algorithmic}
	\label{alg-RW1-WAS-new}
\end{algorithm}

\begin{algorithm}[h!]
\renewcommand{\thealgorithm}{C2}
	\caption{Remove/weaken/add weakened axiom sets one at a time}
     \hspace*{\algorithmicindent}\textbf{Input}: TBox ${\cal T}$,  Oracle Or, set of unwanted axioms $W$ \\
   \hspace*{\algorithmicindent}\textbf{Output}: A repaired TBox
	\begin{algorithmic}[1]
	\State ${\cal T}_r$ $\leftarrow$ ${\cal T}$
	\ForEach {$\alpha$ $\sqsubseteq$ $\beta$ $\in$ $W$}
	\State ${\cal T}_r$ $\leftarrow$ Remove-axioms(${\cal T}_r$, $\{\alpha \sqsubseteq\beta\}$)
   \State $w_{\alpha \sqsubseteq \beta}\leftarrow$ weakened-axiom-set($\alpha \sqsubseteq \beta$, ${\cal T}_r, Or)$
   \State ${\cal T}_r$ $\leftarrow$ Add-axioms(${\cal T}_r$,$w_{\alpha \sqsubseteq \beta}$)
   \EndFor
   \State \Return ${\cal T}_r$
	\end{algorithmic}
	\label{alg-RWWAS1}
	\end{algorithm}

\begin{algorithm}[h!]
\renewcommand{\thealgorithm}{C3}
	\caption{Remove all wrong, weaken all and add weakened axiom sets at end}
     \hspace*{\algorithmicindent}\textbf{Input}: TBox ${\cal T}$,  Oracle Or, set of unwanted axioms $W$ \\
   \hspace*{\algorithmicindent}\textbf{Output}: A repaired TBox
	\begin{algorithmic}[1]
	\State ${\cal T}_r$ $\leftarrow$ Remove-axioms(${\cal T}$, $W$)
	\ForEach {$\alpha$ $\sqsubseteq$ $\beta$ $\in$ $W$}
   \State $w_{\alpha \sqsubseteq \beta}\leftarrow$ weakened-axiom-set($\alpha \sqsubseteq \beta$, ${\cal T}_r, Or)$
   \EndFor
   \State \Return Add-axioms(${\cal T}_r$,$\bigcup_{\alpha \sqsubseteq\beta}$ $w_{\alpha \sqsubseteq \beta}$)
	\end{algorithmic}
	\label{alg-R-W1-WAS}
	\end{algorithm}

\begin{algorithm}[ht!]
\renewcommand{\thealgorithm}{C4}
	\caption{Remove all wrong, weaken/add weakened axiom sets one at a time}
     \hspace*{\algorithmicindent}\textbf{Input}: TBox ${\cal T}$,  Oracle Or, set of unwanted axioms $W$ \\
   \hspace*{\algorithmicindent}\textbf{Output}: A repaired TBox
	\begin{algorithmic}[1]
	\State ${\cal T}_r$ $\leftarrow$ Remove-axioms(${\cal T}$, $W$)
	\ForEach {$\alpha$ $\sqsubseteq$ $\beta$ $\in$ $W$}
   \State $w_{\alpha \sqsubseteq \beta}\leftarrow$ weakened-axiom-set($\alpha \sqsubseteq \beta$, ${\cal T}_r, Or)$
   \State ${\cal T}_r$ $\leftarrow$ Add-axioms(${\cal T}_r$,$w_{\alpha \sqsubseteq \beta}$)
   \EndFor
   \State \Return ${\cal T}_r$
	\end{algorithmic}
	\label{alg-R-WWAS1-new}
	\end{algorithm}

\begin{algorithm}[h!]
\renewcommand{\thealgorithm}{C5}
	\caption{Weaken one at a time, complete one at a time, add completed axiom set and remove all wrong at end}
     \hspace*{\algorithmicindent}\textbf{Input}: TBox ${\cal T}$,  Oracle Or, set of unwanted axioms $W$ \\
   \hspace*{\algorithmicindent}\textbf{Output}: A repaired TBox
	\begin{algorithmic}[1]
	\ForEach {$\alpha$ $\sqsubseteq$ $\beta$ $\in$ $W$}
	\State ${\cal T}_r$ $\leftarrow$ Remove-axioms(${\cal T}$, $\{\alpha \sqsubseteq\beta\}$)
   \State $w_{\alpha \sqsubseteq \beta}\leftarrow$ weakened-axiom-set($\alpha \sqsubseteq \beta$, ${\cal T}_r, Or)$
   \EndFor
   	\ForEach {$\alpha$ $\sqsubseteq$ $\beta$ $\in$ $W$}
   	\State $c_{\alpha \sqsubseteq \beta}\leftarrow$ $\emptyset$
	\ForEach {$sb$ $\sqsubseteq$ $sp$ $\in$ $w_{\alpha \sqsubseteq \beta}$}
   \State $c_{sb \sqsubseteq sp}\leftarrow$ completed-axiom-set($sb \sqsubseteq sp$, ${\cal T}, Or)$
   \State $c_{\alpha \sqsubseteq \beta}\leftarrow$ $c_{\alpha \sqsubseteq \beta}$ $\cup$ $c_{sb \sqsubseteq sp}$
   \EndFor
  \EndFor
  \State ${\cal T}_r$ $\leftarrow$  Add-axioms(${\cal T}$,$\bigcup_{\alpha \sqsubseteq\beta}$ $c_{\alpha \sqsubseteq \beta}$)
   \State \Return Remove-axioms(${\cal T}_r$,$W$)
	\end{algorithmic}
	\label{alg-RW1-C1-CAS}
\end{algorithm}
 
\begin{algorithm}[h!]
\renewcommand{\thealgorithm}{C6}
	\caption{Weaken/complete/add completed axiom sets one at a time, remove all wrong at end} 
     \hspace*{\algorithmicindent}\textbf{Input}: TBox ${\cal T}$,  Oracle Or, set of unwanted axioms $W$ \\
   \hspace*{\algorithmicindent}\textbf{Output}: A repaired TBox
	\begin{algorithmic}[1]
	\ForEach {$\alpha$ $\sqsubseteq$ $\beta$ $\in$ $W$}
	\State ${\cal T}_r$ $\leftarrow$ Remove-axioms(${\cal T}$, $\{\alpha \sqsubseteq\beta\}$)
   \State $w_{\alpha \sqsubseteq \beta}\leftarrow$ weakened-axiom-set($\alpha \sqsubseteq \beta$, ${\cal T}_r, Or)$
    	\State $c_{\alpha \sqsubseteq \beta}\leftarrow$ $\emptyset$
	\ForEach {$sb$ $\sqsubseteq$ $sp$ $\in$ $w_{\alpha \sqsubseteq \beta}$}
   \State $c_{sb \sqsubseteq sp}\leftarrow$ completed-axiom-set($sb \sqsubseteq sp$, ${\cal T}_r, Or)$
   \State ${\cal T}$ $\leftarrow$  Add-axioms(${\cal T}$, $c_{sb \sqsubseteq sp}$)
   \EndFor
   \EndFor
   \State \Return Remove-axioms(${\cal T}$,$W$)
	\end{algorithmic}
	\label{algorithm8_V1}
\end{algorithm}

\begin{algorithm}[!h]
\renewcommand{\thealgorithm}{C7}
	\caption{Remove/weaken/complete/add completed axiom sets one at a time }
     \hspace*{\algorithmicindent}\textbf{Input}: TBox ${\cal T}$,  Oracle Or, set of unwanted axioms $W$ \\
   \hspace*{\algorithmicindent}\textbf{Output}: A repaired TBox
	\begin{algorithmic}[1]
	\State ${\cal T}_r$ $\leftarrow$ ${\cal T}$
	\ForEach {$\alpha$ $\sqsubseteq$ $\beta$ $\in$ $W$}
	\State ${\cal T}_r$ $\leftarrow$ Remove-axioms(${\cal T}_r$, $\{\alpha \sqsubseteq\beta\}$)
   \State $w_{\alpha \sqsubseteq \beta}\leftarrow$ weakened-axiom-set($\alpha \sqsubseteq \beta$, ${\cal T}_r, Or)$
    	\State $c_{\alpha \sqsubseteq \beta}\leftarrow$ $\emptyset$
	\ForEach {$sb$ $\sqsubseteq$ $sp$ $\in$ $w_{\alpha \sqsubseteq \beta}$}
   \State $c_{sb \sqsubseteq sp}\leftarrow$ completed-axiom-set($sb \sqsubseteq sp$, ${\cal T}_r, Or)$
   \State ${\cal T}_r$ $\leftarrow$  Add-axioms(${\cal T}_r$, $c_{sb \sqsubseteq sp}$)
   \EndFor
   \EndFor
   \State \Return${\cal T}_r$
	\end{algorithmic}
	\label{algorithm8_V2}
\end{algorithm}
 
 \begin{algorithm}[ht!]
\renewcommand{\thealgorithm}{C8}
	\caption{Weaken/complete one at a time, add completed axiom sets and remove all wrong axioms at end}
     \hspace*{\algorithmicindent}\textbf{Input}: TBox ${\cal T}$,  Oracle Or, set of unwanted axioms $W$ \\
   \hspace*{\algorithmicindent}\textbf{Output}: A repaired TBox
	\begin{algorithmic}[1]
	\ForEach {$\alpha$ $\sqsubseteq$ $\beta$ $\in$ $W$}
	\State ${\cal T}_r$ $\leftarrow$ Remove-axioms(${\cal T}$, $\{\alpha \sqsubseteq\beta\}$)
   \State $w_{\alpha \sqsubseteq \beta}\leftarrow$ weakened-axiom-set($\alpha \sqsubseteq \beta$, ${\cal T}_r, Or)$
    	\State $c_{\alpha \sqsubseteq \beta}\leftarrow$ $\emptyset$
	\ForEach {$sb$ $\sqsubseteq$ $sp$ $\in$ $w_{\alpha \sqsubseteq \beta}$}
   \State $c_{sb \sqsubseteq sp}\leftarrow$ completed-axiom-set($sb \sqsubseteq sp$, ${\cal T}, Or)$
   \State $c_{\alpha \sqsubseteq \beta}\leftarrow$ $c_{\alpha \sqsubseteq \beta}$ $\cup$ $c_{sb \sqsubseteq sp}$
   \EndFor
   \EndFor
  \State ${\cal T}_r$ $\leftarrow$  Add-axioms(${\cal T}$,$\bigcup_{\alpha \sqsubseteq\beta}$ $c_{\alpha \sqsubseteq \beta}$)
   \State \Return Remove-axioms(${\cal T}_r$,$W$)
	\end{algorithmic}
	\label{alg-RWC1-CAS}
\end{algorithm}

\begin{algorithm}[h!]
\renewcommand{\thealgorithm}{C9}
	\caption{Weaken one at a time, remove all wrong, complete one at a time, then add completed axiom sets at end}
     \hspace*{\algorithmicindent}\textbf{Input}: TBox ${\cal T}$,  Oracle Or, set of unwanted axioms $W$ \\
   \hspace*{\algorithmicindent}\textbf{Output}: A repaired TBox
	\begin{algorithmic}[1]
	\ForEach {$\alpha$ $\sqsubseteq$ $\beta$ $\in$ $W$}
	\State ${\cal T}_r$ $\leftarrow$ Remove-axioms(${\cal T}$, $\{\alpha \sqsubseteq\beta\}$)
   \State $w_{\alpha \sqsubseteq \beta}\leftarrow$ weakened-axiom-set($\alpha \sqsubseteq \beta$, ${\cal T}_r, Or)$
   \EndFor
    \State ${\cal T}_r$ $\leftarrow$ Remove-axioms(${\cal T}_r$,$W$)
   	\ForEach {$\alpha$ $\sqsubseteq$ $\beta$ $\in$ $W$}
   	\State $c_{\alpha \sqsubseteq \beta}\leftarrow$ $\emptyset$
	\ForEach {$sb$ $\sqsubseteq$ $sp$ $\in$ $w_{\alpha \sqsubseteq \beta}$}
   \State $c_{sb \sqsubseteq sp}\leftarrow$ completed-axiom-set($sb \sqsubseteq sp$, ${\cal T}_r, Or)$
   \State $c_{\alpha \sqsubseteq \beta}\leftarrow$ $c_{\alpha \sqsubseteq \beta}$ $\cup$ $c_{sb \sqsubseteq sp}$
   \EndFor
  \EndFor
  \State ${\cal T}_r$ $\leftarrow$  Add-axioms(${\cal T}_r$,$\bigcup_{\alpha \sqsubseteq\beta}$ $c_{\alpha \sqsubseteq \beta}$)
   \State \Return ${\cal T}_r$
	\end{algorithmic}
	\label{algorithm9_V1-new}
\end{algorithm}

\begin{algorithm}[h!]
\renewcommand{\thealgorithm}{C10}
	\caption{Weaken one at a time, remove all wrong, complete/add completed axiom sets one at a time}
     \hspace*{\algorithmicindent}\textbf{Input}: TBox ${\cal T}$,  Oracle Or, set of unwanted axioms $W$ \\
   \hspace*{\algorithmicindent}\textbf{Output}: A repaired TBox
	\begin{algorithmic}[1]
	\ForEach {$\alpha$ $\sqsubseteq$ $\beta$ $\in$ $W$}
	\State ${\cal T}_r$ $\leftarrow$ Remove-axioms(${\cal T}$, $\{\alpha \sqsubseteq\beta\}$)
   \State $w_{\alpha \sqsubseteq \beta}\leftarrow$ weakened-axiom-set($\alpha \sqsubseteq \beta$, ${\cal T}_r, Or)$
   \EndFor
   \State ${\cal T}_r$ $\leftarrow$ Remove-axioms(${\cal T}$, $W$)
   	\ForEach {$\alpha$ $\sqsubseteq$ $\beta$ $\in$ $W$}
	\ForEach {$sb$ $\sqsubseteq$ $sp$ $\in$ $w_{\alpha \sqsubseteq \beta}$}
   \State $c_{sb \sqsubseteq sp}\leftarrow$ completed-axiom-set($sb \sqsubseteq sp$, ${\cal T}_r, Or)$
   \State ${\cal T}_r$ $\leftarrow$  Add-axioms(${\cal T}_r$,$c_{sb \sqsubseteq sp}$)
   \EndFor
  \EndFor
   \State \Return ${\cal T}_r$
	\end{algorithmic}
	\label{alg-RW1-CCAS1-new}
\end{algorithm} 
 
\begin{algorithm}[h!]
\renewcommand{\thealgorithm}{C11}
	\caption{Remove/Weaken one at a time, add the wrong axiom and then complete/add completed axiom sets one at a time, remove all wrong at end}
     \hspace*{\algorithmicindent}\textbf{Input}: TBox ${\cal T}$,  Oracle Or, set of unwanted axioms $W$ \\
   \hspace*{\algorithmicindent}\textbf{Output}: A repaired TBox
	\begin{algorithmic}[1]
	\ForEach {$\alpha$ $\sqsubseteq$ $\beta$ $\in$ $W$}
	\State ${\cal T}_r$ $\leftarrow$ Remove-axioms(${\cal T}$, $\{\alpha \sqsubseteq\beta\}$)
   \State $w_{\alpha \sqsubseteq \beta}\leftarrow$ weakened-axiom-set($\alpha \sqsubseteq \beta$, ${\cal T}_r, Or)$
   \EndFor
   	\ForEach {$\alpha$ $\sqsubseteq$ $\beta$ $\in$ $W$}
   	\State ${\cal T}_r$ $\leftarrow$ Remove-axioms(${\cal T}$, $\{\alpha \sqsubseteq\beta\}$)
	\ForEach {$sb$ $\sqsubseteq$ $sp$ $\in$ $w_{\alpha \sqsubseteq \beta}$}
   \State $c_{sb \sqsubseteq sp}\leftarrow$ completed-axiom-set($sb \sqsubseteq sp$, ${\cal T}_r, Or)$
   \State ${\cal T}$ $\leftarrow$  Add-axioms(${\cal T}$,$c_{sb \sqsubseteq sp}$)
   \EndFor
  \EndFor
   \State \Return Remove-axioms(${\cal T}$,$W$)
	\end{algorithmic}
	\label{Algorithm10_V1}
\end{algorithm}
 
\begin{algorithm}[h!]
\renewcommand{\thealgorithm}{C12}
	\caption{Remove all wrong, weaken all, complete all, add completed axiom sets at end}
     \hspace*{\algorithmicindent}\textbf{Input}: TBox ${\cal T}$,  Oracle Or, set of unwanted axioms $W$ \\
   \hspace*{\algorithmicindent}\textbf{Output}: A repaired TBox
	\begin{algorithmic}[1]
	\State ${\cal T}_r$ $\leftarrow$ Remove-axioms(${\cal T}$, $W$)
	\ForEach {$\alpha$ $\sqsubseteq$ $\beta$ $\in$ $W$}
   \State $w_{\alpha \sqsubseteq \beta}\leftarrow$ weakened-axiom-set($\alpha \sqsubseteq \beta$, ${\cal T}_r, Or)$
   \EndFor
    \ForEach {$\alpha$ $\sqsubseteq$ $\beta$ $\in$ $W$}
       \State $c_{\alpha \sqsubseteq \beta}\leftarrow$ $\emptyset$
         \ForEach {$sb$ $\sqsubseteq$ $sp$ $\in$ $w_{\alpha \sqsubseteq \beta}$}
       \State $c_{sb \sqsubseteq sp}\leftarrow$ completed-axiom-set($sb \sqsubseteq sp$, ${\cal T}_r, Or)$
       \State $c_{\alpha \sqsubseteq \beta}\leftarrow$ $c_{\alpha \sqsubseteq \beta}$ $\cup$ $c_{sb \sqsubseteq sp}$
   \EndFor
   \EndFor
   \State \Return Add-axioms(${\cal T}_r$,$\bigcup_{\alpha \sqsubseteq\beta}$ $c_{\alpha \sqsubseteq \beta}$)
	\end{algorithmic}
	\label{alg-R-W-C-CAS}
	\end{algorithm}

\begin{algorithm}[h!]
\renewcommand{\thealgorithm}{C13}
	\caption{Remove all wrong, weaken/complete/add completed axiom sets one at a time}
     \hspace*{\algorithmicindent}\textbf{Input}: TBox ${\cal T}$,  Oracle Or, set of unwanted axioms $W$ \\
   \hspace*{\algorithmicindent}\textbf{Output}: A repaired TBox
	\begin{algorithmic}[1]
	\State ${\cal T}_r$ $\leftarrow$ Remove-axioms(${\cal T}$, $W$)
	\ForEach {$\alpha$ $\sqsubseteq$ $\beta$ $\in$ $W$}
   \State $w_{\alpha \sqsubseteq \beta}\leftarrow$ weakened-axiom-set($\alpha \sqsubseteq \beta$, ${\cal T}_r, Or)$
   	\ForEach {$sb$ $\sqsubseteq$ $sp$ $\in$ $w_{\alpha \sqsubseteq \beta}$}
   \State $c_{sb \sqsubseteq sp}\leftarrow$ completed-axiom-set($sb \sqsubseteq sp$, ${\cal T}_r, Or)$
   \State ${\cal T}_r$ $\leftarrow$  Add-axioms(${\cal T}_r$,$c_{sb \sqsubseteq sp}$)
   \EndFor
   \EndFor
   \State \Return ${\cal T}_r$
	\end{algorithmic}
	\label{alg-R-WCCAS1}
	\end{algorithm}

\begin{table*}[!h]
\begin{center}
\caption{\label{tab:algorithm-overview}Algorithms.}
\begin{tabular}{|l|l|}
\hline
Algorithm & Description \\
\hline
C1 & Weaken one at a time, add weakened axiom sets and remove all wrong at end\\
C2 & Remove/weaken/add weakened axiom sets one at a time \\
C3 & Remove all wrong, weaken one at a time, add weakened axiom sets at end\\
C4 & Remove all wrong, weaken/add weakened axiom sets one at a time \\
C5 & Weaken one at a time, complete one at a time, add completed axiom sets and remove all\\& wrong at end \\
C6 & Weaken/complete/add completed axiom sets one at a time, remove all wrong at end\\
C7 & Remove/weaken/complete/add completed axiom sets one at a time\\
C8 & Weaken/complete one at a time, add completed axiom sets and remove all wrong at end\\
C9 &Weaken one at a time, remove all wrong, complete one at a time, then add\\& completed axiom sets at end\\
C10 & Weaken one at a time, remove all wrong, complete/add completed axiom sets one at a time\\
C11 & Weaken one at a time, complete/add completed axiom sets one at a time, remove all wrong\\& at end\\
C12 & Remove all wrong, weaken all, complete all, add completed axiom sets at end \\
C13 & Remove all wrong, weaken/complete/add completed axiom sets one at a time \\
\hline

\end{tabular}
\end{center}
\end{table*}

\clearpage

\section*{Appendix B - Experiments}

In this appendix, we give the full results of the comparative experiments run in different ontologies

Table \ref{tab:wrong axioms} lists the wrong axioms we introduced in each test ontology for experiments. These wrong axioms were generated by replacing existing axioms with axioms where their left/right-hand side concepts were changed. Figure \ref{fig-Mini-GALEN-viz} visualizes the structure of the Mini-Galen ontology in Figure \ref{fig-Mini-GALEN}. 

The full results of the experiments are listed in Tables \ref{tab:comparison2}-\ref{tab:completing comparison16}. 
Table \ref{tab:comparison-order} shows the sizes of the sub-and super-concepts sets for weakening when removing wrong axioms one at a time in different orders using Algorithm \ref{alg-RWWAS1}.
Tables \ref{tab:comparison2}-\ref{tab:comparison6} show the sizes of the super- and sub-concepts sets for weakening different ontologies using Algorithms \ref{alg-RW1-WAS}-\ref{alg-R-WWAS1}. 

When the completion is added, first, the Mini-GALEN ontology and NCI ontology are employed to run several comparative experiments showing the difference between the sizes of source/target sets and the sizes of Sup($\alpha$,${\cal T}$)/Sub($\beta$,${\cal T}$) sets. Tables \ref{tab:completing comparison1 }-\ref{tab:completing comparison12} list the relevant completing results using Algorithms \ref{alg-RW1-C1-CAS}-\ref{alg-R-WCCAS1}. For the remaining ontologies,  in order to not introduce equivalence between concepts in the ontology, we only choose the concepts in the source and target sets to generate the completed axioms and Tables \ref{tab:completing comparison5}-\ref{tab:completing comparison16} show the results of the sizes of the source and target sets when completing different ontologies using Algorithms \ref{alg-RW1-C1-CAS}-\ref{alg-R-WCCAS1}.

\begin{table*}[h!]
\begin{center}
    
\caption{\label{tab:wrong axioms}Wrong axioms in each ontology.}
\begin{tabular}{|p{1.5cm}|p{11.3cm}|}
\hline
Ontology&Wrong axioms\\\hline
Mini-GALEN&PathologicalProcess$\sqsubseteq$InflammationProcess,

InflammationProcess$\sqsubseteq$GranulomaProcess, 

Endocarditis$\sqsubseteq$PathologicalProcess\\\hline
PACO&Polish\_car$\sqsubseteq$Home\_improvement\_maintenance,  

Washing\_windows$\sqsubseteq$Home\_improvement\_maintenance,

Moderate$\sqsubseteq$Speed,
Washing\_car$\sqsubseteq$Home\_improvement\_maintenance,

Walking$\sqsubseteq$Daily\_living\_activity,   
Per\_week$\sqsubseteq$By\_duration \\\hline
EKAW&Camera\_Ready\_Paper$\sqsubseteq\exists$writtenBy.Student,
Tutorial$\sqsubseteq$Conference, 

Invited\_Talk\_Abstract$\sqsubseteq$Paper, 
Programme\_Brochure$\sqsubseteq$Flyer\\\hline
NICI&Tooth\_tissue$\sqsubseteq$Tooth, Red\_fiber$\sqsubseteq$Connective\_tissue\_fiber, Eye\_lid$\sqsubseteq$Cheek\\\hline
Pizza&PineKernels$\sqsubseteq$VegetableTopping, PeperoniSausageTopping$\sqsubseteq$PeperonataTopping,

IceCream$\sqsubseteq\exists$hasTopping.FruitTopping, 
RosemaryTopping$\sqsubseteq$VegetableTopping \\\hline
OFSMR&Beverage$\sqsubseteq$Food, 
Bread$\sqsubseteq$Procesed\_fruit\_and\_vegetables,

Pasta$\sqsubseteq$Procesed\_fruit\_and\_vegetables
\\\hline
\end{tabular}
\end{center}

\end{table*}

\begin{table*}[h!]
\begin{center}
\caption{\label{tab:comparison-order}Removing wrong axioms in different order for Mini-GALEN by Algorithm \ref{alg-RWWAS1}. Wrong axioms: \textcircled{1}PPr$\subseteq$IPr, \textcircled{2}IPr$\subseteq$GPr, \textcircled{3}E$\subseteq$PPr.}
\begin{tabular}{|p{1.955cm}|p{1.65cm}|p{1.65cm}|p{1.65cm}|p{1.65cm}|p{1.65cm}|p{1.65cm}|}
\hline
Wrong Axiom&\textcircled{1}$\rightarrow$\textcircled{2}$\rightarrow$\textcircled{3}&\textcircled{1}$\rightarrow$\textcircled{3}$\rightarrow$\textcircled{2}&\textcircled{2}$\rightarrow$\textcircled{1}$\rightarrow$\textcircled{3} &\textcircled{2}$\rightarrow$\textcircled{3}$\rightarrow$\textcircled{1} &\textcircled{3}$\rightarrow$\textcircled{2}$\rightarrow$\textcircled{1}&\textcircled{3}$\rightarrow$\textcircled{1}$\rightarrow$\textcircled{2} \\\hline
Sup($\beta$,${\cal T}$)&3  2  2&3  2  2& 3  2 2&2 2 3&2  2  4&3 2 4  \\
Sub($\alpha$,${\cal T}$)&2 1 1&2 1 1&2 1 3&1 3 1&1 2 1&1 1 1\\\hline
\end{tabular}
\end{center}

\end{table*}

\begin{table*}[h!]
\begin{center}

\caption{\label{tab:comparison2}Weakening the PACO ontology using Algorithms C1-C4. Six wrong axioms give 6 sup/sub-sets per algorithm.}
\begin{tabular}{|l|r|r|r|r|r|}
\hline
 &C1&C2&C3&C4\\\hline
 Sup($\beta$,${\cal T}$)&4 4 4 3 4 3&4 4 4 3 4 3&4 4 4 3 4 3&4 4 4 3 4 3\\
 Sub($\alpha$,${\cal T}$)&1 1 1 6 1 1&1 1 1 6 1 1&1 1 1 6 1 1&1 1 1 6 1 1\\\hline
\end{tabular}
\end{center}

\end{table*}

\begin{table*}[h!]
\begin{center}
\caption{\label{tab:comparison3}Weakening the EKAW ontology using Algorithms C1-C4. Four wrong axioms give 4 sup/sub-sets per algorithm.}
\begin{tabular}{|l|r|r|r|r|}
\hline
 &C1&C2&C3&C4\\\hline
 Sup($\beta$,${\cal T}$)&3 4 3 3&3 4 3 3&3 4 3 3&3 4 3 3\\
 Sub($\alpha$,${\cal T}$)&1 1 1 1&1 1 1 1&1 1 1 1&1 1 1 1\\\hline
\end{tabular}
\end{center}

\end{table*}

\begin{table*}[h!]
\begin{center}
\caption{\label{tab:comparison4}Weakening the NCI ontology using Algorithms C1-C4. Three wrong axioms give 3 sup/sub-sets per algorithm.}
\begin{tabular}{|l|r|r|r|r|r|}
\hline
&C1&C2&C3&C4\\\hline
 Sup($\beta$,${\cal T}$)&13 15 8&13 15 8&13 15 8&13 15 8\\
 Sub($\alpha$,${\cal T}$)& 7 1 3& 7 1 3& 7 1 3& 7 1 3\\\hline
\end{tabular}
\end{center}

\end{table*}

\begin{table*}[h!]
\begin{center}
\caption{\label{tab:comparison5}Weakening the Pizza ontology using Algorithms C1-C4. Four wrong axioms give 4 sup/sub-sets per algorithm.}
\begin{tabular}{|l|r|r|r|r|r|}
\hline
 &C1&C2&C3&C4\\\hline
 Sup($\beta$,${\cal T}$)&4 8 4 8&4 8 4 8&4 8 4 8&4 8 4 8\\
 Sub($\alpha$,${\cal T}$)&1 1 1 1&1 1 1 1&1 1 1 1&1 1 1 1\\\hline
\end{tabular}
\end{center}

\end{table*}

\begin{table*}[h!]
\begin{center}
\caption{\label{tab:comparison6}Weakening the OFSMR ontology using Algorithms C1-C4. Three wrong axioms give 3 sup/sub-sets per algorithm.}
\begin{tabular}{|l|r|r|r|r|r|}
\hline
 &C1&C2&C3&C4\\\hline
 Sup($\beta$,${\cal T}$)&2 4 4 &2 4 4 &2 4 4 &2 4 4 \\\hline
 Sub($\alpha$,${\cal T}$)&2 1 1&2 1 1&2 1 1&2 1 1\\\hline
\end{tabular}
\end{center}

\end{table*}

\begin{table*}[!h]
\begin{center}
    
\caption{\label{tab:completing comparison1 }Completing the Mini-GALEN ontology using Algorithms C5-C7.}
\begin{tabular}{|l|r|r|r|r|r|r|}
\hline
 &C5&C6&C7\\\hline
 Weakened&PPr$\sqsubseteq$NPr, IPr$\sqsubseteq$NPr
&PPr$\sqsubseteq$NPr, IPr$\sqsubseteq$NPr
&PPr$\sqsubseteq$NPr, IPr$\sqsubseteq$NPr
\\\hline
 Source &1 1&1 1&1 1\\
 Target &3 2&3 2&3 4\\\hline
 Completed&PPr$\sqsubseteq$NPr, IPr$\sqsubseteq$NPr
 &PPr$\sqsubseteq$NPr, IPr$\sqsubseteq$NPr
 &PPr$\sqsubseteq$NPr, IPr$\sqsubseteq$PPr\\\hline
\end{tabular}
\end{center}

\end{table*}

\begin{table*}[!h]
\begin{center}
\caption{\label{tab:completing comparison2 }Completing the Mini-GALEN ontology using Algorithms C5-C7.}

\begin{tabular}{|l|r|r|r|r|r|r|}
\hline
 &C5&C6&C7\\\hline
 Weakened
&PPr$\sqsubseteq$NPr, IPr$\sqsubseteq$NPr
&PPr$\sqsubseteq$NPr, IPr$\sqsubseteq$NPr
&PPr$\sqsubseteq$NPr, IPr$\sqsubseteq$NPr\\\hline
 Sup($\alpha$,${\cal T}$) &1 1&1 1&1 1\\
 Sub($\beta$,${\cal T}$) &3 2&3 4&3 4\\\hline
 Completed
 &PPr$\sqsubseteq$NPr, IPr$\sqsubseteq$NPr
 &PPr$\sqsubseteq$NPr, IPr$\sqsubseteq$PPr
 &PPr$\sqsubseteq$NPr, IPr$\sqsubseteq$PPr\\\hline
\end{tabular}
\end{center}

\end{table*}

\begin{table*}[!h]
\begin{center}
\caption{\label{tab:completing comparison3 }Completing the Mini-GALEN ontology using Algorithms C8-C13.}
\begin{tabular}{|p{1.5cm}|p{1.45cm}|p{1.45cm}|p{1.45cm}|p{1.45cm}|p{1.45cm}|p{1.45cm}|}
\hline
 &C8&C9&C10&C11&C12&C13\\\hline
 Weakened&PPr$\sqsubseteq$NPr, IPr$\sqsubseteq$NPr
&PPr$\sqsubseteq$NPr, IPr$\sqsubseteq$NPr
&PPr$\sqsubseteq$NPr, IPr$\sqsubseteq$NPr
&PPr$\sqsubseteq$NPr, IPr$\sqsubseteq$NPr
&IPr$\sqsubseteq$NPr
&IPr$\sqsubseteq$NPr\\\hline
Source &3 2&1 1&1 1&1 1&1 &1\\
Target &3 2&2 2&2 3&3 2&2 &2\\\hline
Completed&GPr$\sqsubseteq$IPr,
PPr$\sqsubseteq$NPr, IPr$\sqsubseteq$NPr
&PPr$\sqsubseteq$NPr, IPr$\sqsubseteq$NPr
&PPr$\sqsubseteq$NPr, IPr$\sqsubseteq$PPr
&PPr$\sqsubseteq$NPr, IPr$\sqsubseteq$NPr
&IPr$\sqsubseteq$NPr&IPr$\sqsubseteq$NPr
\\\hline
\end{tabular}    
\end{center}

\end{table*}

\begin{table*}[!h]
\begin{center}
\caption{\label{tab:completing comparison4 }Completing the Mini-GALEN ontology using Algorithms C8-C13.}
\begin{tabular}{|p{1.5cm}|p{1.45cm}|p{1.45cm}|p{1.45cm}|p{1.45cm}|p{1.45cm}|p{1.45cm}|}
\hline
&C8&C9&C10&C11&C12&C13\\\hline
 Weakened&PPr$\sqsubseteq$NPr, IPr$\sqsubseteq$NPr
&PPr$\sqsubseteq$NPr, IPr$\sqsubseteq$NPr
&PPr$\sqsubseteq$NPr, IPr$\sqsubseteq$NPr
&PPr$\sqsubseteq$NPr, IPr$\sqsubseteq$NPr
&IPr$\sqsubseteq$NPr
&IPr$\sqsubseteq$NPr\\\hline
Sup($\alpha$,${\cal T}$) &4 3&1 1&1 1&1 1&1 &1\\
Sub($\beta$,${\cal T}$) &5 5&2 2&2 3&3 4&2 &2\\\hline
Completed&GPr$\sqsubseteq$IPr,
PPr$\sqsubseteq$NPr, IPr$\sqsubseteq$PPr
&PPr$\sqsubseteq$NPr, IPr$\sqsubseteq$NPr
&PPr$\sqsubseteq$NPr, IPr$\sqsubseteq$PPr
&PPr$\sqsubseteq$NPr, IPr$\sqsubseteq$PPr
&IPr$\sqsubseteq$NPr&IPr$\sqsubseteq$NPr
\\\hline
\end{tabular}    
\end{center}

\end{table*}

\begin{table*}[!h]
\begin{center}
\caption{\label{tab:completing comparison9}Completing the NCI ontology using Algorithms C5-C9.}

\begin{tabular}{|l|r|r|r|r|r|r|}
\hline
 &C5&C6&C7&C8&C9\\\hline
 Source&3 1 1&3 1 1&3 1 1&6 14 6&3 1 1\\
 Target&40 2143 83&40 2143 83&40 2136 76&59 2143 83&40 2133 76\\\hline
\end{tabular}    
\end{center}
\end{table*}

\begin{table*}[!h]
\begin{center}
\caption{\label{tab:completing comparison10}Completing the NCI ontology using Algorithms C5-C9.}

\begin{tabular}{|l|r|r|r|r|r|r|}
\hline
 &C5&C6&C7&C8&C9\\\hline
 Sup($\alpha$,${\cal T}$)&3 1 1&3 1 1&3 1 1&15 16 9&3 1 1\\
 Sub($\beta$,${\cal T}$)&41 2143 83&41 2143 83&41 2136 76&66 2144 86&41 2133 76\\\hline
\end{tabular}    
\end{center}
\end{table*}

\begin{table*}[!h]
\begin{center}
\caption{\label{tab:completing comparison11}Completing the NCI ontology using Algorithms C10-C13.}
\begin{tabular}{|l|r|r|r|r|r|r|}
\hline
 &C10&C11&C12&C13\\\hline
 Source&3 1 1&3 1 1&3 1 1&3 1 1\\
 Target&40 2136 76&40 2143 83&40 2133 76&40 2133 76\\\hline
\end{tabular}    
\end{center}

\end{table*}

\begin{table*}[!h]
\centering
\caption{\label{tab:completing comparison12}Completing the NCI ontology using Algorithms C10-C13.}
\begin{tabular}{|l|r|r|r|r|r|r|}
\hline
 &C10&C11&C12&C13\\\hline
 Sup($\alpha$,${\cal T}$)&3 1 1&3 1 1&3 1 1&3 1 1\\
 Sub($\beta$,${\cal T}$)&41 2136 76&41 2143 83&41 2133 76&41 2133 76\\\hline
\end{tabular}

\end{table*}

\begin{table*}[!h]
\begin{center}
 
\caption{\label{tab:completing comparison5}Completing the PACO ontology using Algorithms C5-C9.}

\begin{tabular}{|l|r|r|r|r|r|}
\hline
 &C5&C6&C7&C8&C9\\\hline
 Source&1 1 1 1 1 1&1 1 1 1 1 1&1 1 1 1 1 1&2 2 2 2 2 3&1 1 1 1 1 1\\
 Target&59 59 59 171 40 40&59 59 59 171 40 40&59 59 59 171 40 40&59 59 59 171 40 40&51 51 51 168 39 39\\\hline
\end{tabular}   
\end{center}
\end{table*}

\begin{table*}[!h]
\begin{center}

\caption{\label{tab:completing comparison6}Completing the PACO ontology using Algorithms C10-C13.}
\begin{tabular}{|l|r|r|r|r|r|}
\hline
 &C10&C11&C12&C13\\\hline
 Source&1 1 1 1 1 1&1 1 1 1 1 1&1 1 1 1 1 1&1 1 1 1 1 1\\
 Target&51 52 53 170 39 40&59 59 59 171 40 40&51 51 51 168 39 39&51 52 53 170 39 40\\\hline
\end{tabular}    
\end{center}

\end{table*}

\begin{table*}[!h]
\begin{center}

\caption{\label{tab:completing comparison7}Completing of the EKAW ontology using Algorithms C5-C9.}
\begin{tabular}{|l|r|r|r|r|r|}
\hline
 &C5&C6&C7&C8&C9\\\hline
 Source&9 1 1 1&9 1 1 1&9 1 1 1&10 2 2 2&9 1 1 1\\
 Target&23 17 34 34&23 17 34 34&23 17 34 34&23 17 34 34&23 17 33 33\\\hline
\end{tabular}    
\end{center}

\end{table*}

\begin{table*}[!h]
\begin{center}

\caption{\label{tab:completing comparison8}Completing of the EKAW ontology using Algorithms C10-C13.}
\begin{tabular}{|l|r|r|r|r|r|}
\hline
 &C10&C11&C12&C13\\\hline
 Source&9 1 1 1&9 1 1 1&9 1 1 1&9 1 1 1\\
 Target&23 17 33 34&23 17 34 34&23 17 33 33&23 17 33 34\\\hline
\end{tabular}    
\end{center}

\end{table*}

\begin{table*}[!h]
\begin{center}

\caption{\label{tab:completing comparison13}Completing by the Pizza ontology using Algorithms C5-C9.}
\begin{tabular}{|l|r|r|r|r|r|}
\hline
 &C5&C6&C7&C8&C9\\\hline
 Source&1 1 3 3&1 1 3 3&1 1 3 3&2 7 4 6&1 1 3 3\\
 Target&50 147 50 50&50 147 50 50&50 147 50 50&50 147 50 50&50 144 48 48\\\hline
\end{tabular}    
\end{center}

\end{table*}

\begin{table*}[!h]
\begin{center}

\caption{\label{tab:completing comparison14}Completing the Pizza ontology using Algorithms C10-C13.}
\begin{tabular}{|l|r|r|r|r|r|}
\hline
 &C10&C11&C12&C13\\\hline
 Source&1 1 3 3&1 1 3 3&1 1 3 3&1 1 3 3\\
 Target&49 147 48 50 &50 147 50 50&18 144 48 48&48 145 49 50\\\hline
\end{tabular}
    
\end{center}

\end{table*}

\begin{table*}[h!]
\begin{center}

\caption{\label{tab:completing comparison15}Completing the OFSMR ontology using Algorithms C5-C9.}
\begin{tabular}{|l|r|r|r|r|r|}
\hline
 &C5&C6&C7&C8&C9\\\hline
 Source&1 1 1&1 1 1&1 1 1&2 3 3&1 1 1\\
 Target&125 125 125&125 125 125&125 125 125&125 125 125&123 122 122\\\hline
\end{tabular}
    
\end{center}

\end{table*}

\begin{table*}[h!]
\begin{center}

\caption{\label{tab:completing comparison16}Completing the OFSMR ontology using Algorithms C10-C13.}
\begin{tabular}{|l|r|r|r|r|r|}
\hline
 &C10&C11&C12&C13\\\hline
 Source&1 1 1&1 1 1&1 1 1&1 1 1\\
 Target&123 122 123&125 125 125&123 122 122&123 122 123\\\hline
\end{tabular}
    
\end{center}

\end{table*}

\clearpage

\section*{Appendix C - Derivation of the Hasse diagrams}

For a given TBox $\mathcal{T}$, let Der($\mathcal{T}$) denote the set of derivable axioms from $\mathcal{T}$.
Then, for TBoxes $\mathcal{T}_1$ and $\mathcal{T}_2$, if $\mathcal{T}_1$ $\sqsubseteq$ $\mathcal{T}_2$, then we know that  Der($\mathcal{T}_1$) $\sqsubseteq$ Der($\mathcal{T}_2$). This means that if an axiom is derivable from TBox $\mathcal{T}_1$, it is also derivable from TBox $\mathcal{T}_2$ (but not necessarily the other way around). As the sub- and super-concepts of a concept are computed using subsumption axioms, this also means that the set of sub-concepts for a concept in $\mathcal{T}_1$ is a subset of the set of sub-concepts for a concept in $\mathcal{T}_2$, and the set of super-concepts for a concept in $\mathcal{T}_1$ is a subset of the set of super-concepts for a concept in $\mathcal{T}_2$.
When computing weakened axiom sets and completed axiom sets, the algorithms compute sets of sub-concepts and sets of super-concepts to generate candidate axioms for these weakened and completed axiom sets. Therefore, if $\mathcal{T}_1$ $\sqsubseteq$ $\mathcal{T}_2$, the sets of candidate axioms for the weakened and completed axiom sets computed for $\mathcal{T}_1$ are subsets of those computed for $\mathcal{T}_2$. This means more validation work for $\mathcal{T}_2$, but also possibly a more complete final ontology.
The Hasse diagrams are based on this observation.

{\it Removing.} In general, when removing all axioms at once, the TBox is a subset of the TBox with one axiom removed, which in turn is a subset of the TBox where no axioms are removed. When adding no axioms back, the TBox is a subset of the TBox with one axiom added back, which in turn is a subset of the TBox where all axioms are added back.
If no wrong axioms are removed, then nothing needs to be added back and thus AB-one, AB-all and AB-none have the same result ($\mathcal{T}_{R-none,AB-all}$ = $\mathcal{T}_{R-none,AB-one}$ = $\mathcal{T}_{R-none,AB-none}$). The TBox for these strategies is larger during computation (of weakened or completed axiom sets) than the TBoxes where one or all wrong axioms are removed.
If one wrong axiom at the time is removed, the adding back all (AB-all) or one (AB-one) give the same result ($\mathcal{T}_{R-one,AB-all}$ = $\mathcal{T}_{R-one,AB-one}$) as both strategies add the same one axiom back. The TBox for these strategies is larger than when no wrong axiom is added back ($\mathcal{T}_{R-one,AB-none}$ $\sqsubseteq$ $\mathcal{T}_{R-one,AB-all}$ = $\mathcal{T}_{R-one,AB-one}$).
When all wrong axioms are removed at once, then they will be added back at the end or not.\footnote{After completing they should be removed, but after weakening they could be added back for the completion step.} However, this does not influence the TBox during the computation. Therefore, the add back strategy does not matter and the TBox during computation is smaller than when wrong axioms were removed one at a time ($\mathcal{T}_{R-all,AB-all}$ = $\mathcal{T}_{R-all,AB-one}$ = $\mathcal{T}_{R-all,AB-none}$  $\sqsubseteq$ $\mathcal{T}_{R-one,AB-none}$).

{\it Weakening.} First, we note that updating immediately or updating after each wrong axiom is the same operation for weakening, as a complete weakened axiom set for a wrong axiom is computed. Thus, the TBox for ($\mathcal{T}_{W-one,U-now}$) is the same as for ($\mathcal{T}_{W-one,U-end\_one}$), and  the TBox for ($\mathcal{T}_{W-all,U-now}$) is the same as for ($\mathcal{T}_{W-all,U-end\_one}$).  Further, when weakening one axiom at a time and updating the TBox (i.e., adding the axioms of the weakened axiom set for a wrong axiom) immediately, results in a larger TBox for the next computations of weakened axiom sets for wrong axioms, than if we would not update immediately ($\mathcal{T}_{W-one,U-end\_all}$ $\sqsubseteq$ $\mathcal{T}_{W-one,U-now}$). When not immediately updating, the TBox for generating the weakened axioms sets stays the same for all wrong axioms and thus gives the same result as weakening all wrong axioms at once. Thus, $\mathcal{T}_{W-all,U-now}$ = $\mathcal{T}_{W-all,U-end\_all}$ = $\mathcal{T}_{W-one,U-end\_all}$.

{\it Completing.} When completing one axiom at a time and updating the TBox (i.e., adding the axioms of the completed axiom set for a weakened axiom) immediately, results in a larger TBox for the next computations of completed axiom sets for weakened axioms than not updating immediately ($\mathcal{T}_{C-one,U-end\_one}$ $\sqsubseteq$ $\mathcal{T}_{C-one,C-now}$, $\mathcal{T}_{C-one,U-end\_all}$ $\sqsubseteq$ $\mathcal{T}_{C-one,C-now}$,). When not updating immediately, there is the choice between updating after all weakened axioms for a particular wrong axiom have been processed or waiting until all weakened axioms for all wrong axioms are processed. The TBox for the former case is larger than the one for the latter case ($\mathcal{T}_{C-one,U-end\_all}$ $\sqsubseteq$ $\mathcal{T}_{C-one,U-end\_one}$). Waiting to update the TBox until all weakened axioms for all wrong axioms are processed, means the TBox stays the same during the computation of the completed axioms sets and thus gives the same result as completing all weakened axioms at once ($\mathcal{T}_{C-one,U-end\_all}$ = $\mathcal{T}_{C-all,U-end\_all}$ = $\mathcal{T}_{C-all,U-end\_one}$ = $\mathcal{T}_{C-all,U-now}$).

\end{document}